\documentclass[11pt,letterpaper]{article}
\usepackage[top=0.6in,left=1.5in,footskip=0.75in,marginparwidth=2in]{geometry}
\date{}

\usepackage[pdftex]{graphicx}
\usepackage{subfigure}

\usepackage{algorithmic}
\usepackage{algorithm}
\usepackage{amssymb,amsmath}
\usepackage{stmaryrd}
\usepackage{xcolor}
\usepackage{multicol}
\usepackage{multirow}
\usepackage[T1]{fontenc}
\usepackage{xintexpr}
\usepackage{leftidx}

\usepackage{rotating}
\usepackage{multirow}
\usepackage{longtable}
\usepackage{colortbl}
\usepackage{dsfont}

\usepackage{pgf}

\usepackage{url}
\DeclareMathOperator*{\argmax}{arg\,max}

\newcommand{\ue}{%
	\begin{tikzpicture}%
	\draw[fill = black] (.25ex,.25ex) circle (.3ex);
	\draw[thick] (.55ex,.25ex) -- (1.55ex,.25ex);%
	\draw[fill = black] (1.85ex, .25ex) circle (.3ex);%
	\end{tikzpicture}%
}

\newcommand{\simpcir}{%
	\begin{tikzpicture}%
	\draw[fill = black] (.25ex,.25ex) circle (.3ex);
	\end{tikzpicture}%
}

\usepackage{tikz}
\usepackage{centernot}%
\usetikzlibrary{shapes.geometric}

\tikzstyle{extremely densely dashed} = [dash pattern=on 2pt off 1pt]
\tikzstyle{pattern}=[draw,circle,black,bottom color=white, top color= white, text=black,minimum width=10pt]
\tikzstyle{peers}=[draw,circle,violet,bottom color=green!60, top color= white, text=violet,minimum width=10pt, scale = 0.7]
\definecolor{burntorange}{cmyk}{0,0.52,1,0}
\def\oran{orange!30}
\tikzstyle{cliquepeers}=[draw, red, fill= red, text=black,minimum width=1pt, scale=0.4]
\tikzstyle{myFivePoly} =  [regular polygon,regular polygon sides=5,minimum width=1pt, scale=0.9]
\tikzstyle{mytriangle} =  [fill=blue!20, regular polygon, regular polygon sides=3,minimum width=1pt, scale=0.4]
\tikzstyle{extremely densely dashed} = [dash pattern=on 2pt off 1pt]
\tikzstyle{finedashpath}=[violet, extremely densely dashed,thick]
\tikzstyle{superpeers}=[draw,circle,thick,burntorange, left color=\oran,text=black, minimum width=20pt]
\tikzstyle{pattern}=[draw,circle,black,bottom color=white, top color= white, text=black,minimum width=10pt]

\newcommand{\samplePattern}{%
	\begin{tikzpicture}%
	\node[pattern, inner sep=0pt, bottom color = gray, scale=0.5] (samplePattern) at (0,0) {};
	\end{tikzpicture}%
}

\newcommand{\inlineimage}[1]{$\vcenter{\hbox{\protect\includegraphics[height=0.8\baselineskip,origin=c]{#1}}}$}

\title{\bf Conceptualization of Object Compositions\\ Using Persistent Homology}

\author{Christian A. Mueller and Andreas Birk %
\thanks{The authors are with the Robotics Group of the Computer Science \& Electrical Engineering Department, Jacobs University Bremen, Germany, \tt \{chr.mueller,a.birk\}@jacobs-university.de } 
}
\begin{document}
\maketitle

\begin{abstract}
  A topological shape analysis is proposed and utilized to learn concepts that reflect shape commonalities.
  Our approach is two-fold: i) a spatial topology analysis of point cloud segment constellations within objects.
  Therein constellations are decomposed and described in an hierarchical manner -- from single segments to segment groups until a single group reflects an entire object.
  ii) a topology analysis of the description space in which segment decompositions are exposed in. 
  Inspired by Persistent Homology, hidden groups of shape commonalities are revealed from object segment decompositions.
  Experiments show that extracted persistent groups of commonalities can represent semantically meaningful shape concepts. 
  We also show the generalization capability of the proposed approach considering samples of external datasets.
\end{abstract}

\section{Introduction}\label{sec:intro}
Reasoning about shapes where \emph{commonalities} lead to similar behavior finds its application in many \emph{robotic areas} ranging from household to industry as in object shape categorization tasks~\cite{6942984}, in generation of grasping primitives for similar object appearances in manipulation~\cite{6696928}, finding substitutes for currently absent objects~\cite{DBLP:conf/icra/AbelhaGS16}, etc.

In object \emph{instance} or \emph{category} recognition tasks, a label is associated to a \emph{specific} instance like \emph{John's mug} or to a \emph{generic} group of instances like \emph{mug} which share commonalities in appearance~\cite{Sloutsky2010}; 
this group of instances can be denoted as \emph{category} whereas the description and abstraction of group commonalities as \emph{concept}.
Associations are generally human-made, individually and continuously evolved over lifetime experience based on a set of modalities like \emph{tactual}, \emph{auditory} or \emph{visual} sensations~\cite{Palmeri2004a}.
The combination of those sensations allows us to reliably interpret perceived object information~\cite{ZimgrodHommel2013}.
Humans are capable of incorporating further modalities like \emph{functional object knowledge} to differentiate even though visual percepts are similar as \emph{mug}, \emph{cup}, \emph{vase} or \emph{bowl}.
From machine-vision perspective, particularly human-supervised learning methods are highly vulnerable to incorporate such knowledge (e.g. function) which is not inferable from the given data (e.g. images or point clouds). 
This is often inevitable when a supervised labeling process is conducted by humans which will ultimately lead to such \emph{biases} in the training phase.
Consequently, by avoiding supervision, a method is proposed that objectively in an unsupervised and data-driven manner learns shape concepts from point clouds irrespective of human-annotations which may contain \emph{biases}. %
Therein, extracted segment constellations within object point clouds are exploited to learn patterns and eventually concepts of shape commonalties in a hierarchical manner.

\section{Motivation and Related Work}\label{sec:rw}
\begin{figure}
 	\small
 	\centering
 	\includegraphics[width=0.999\linewidth]{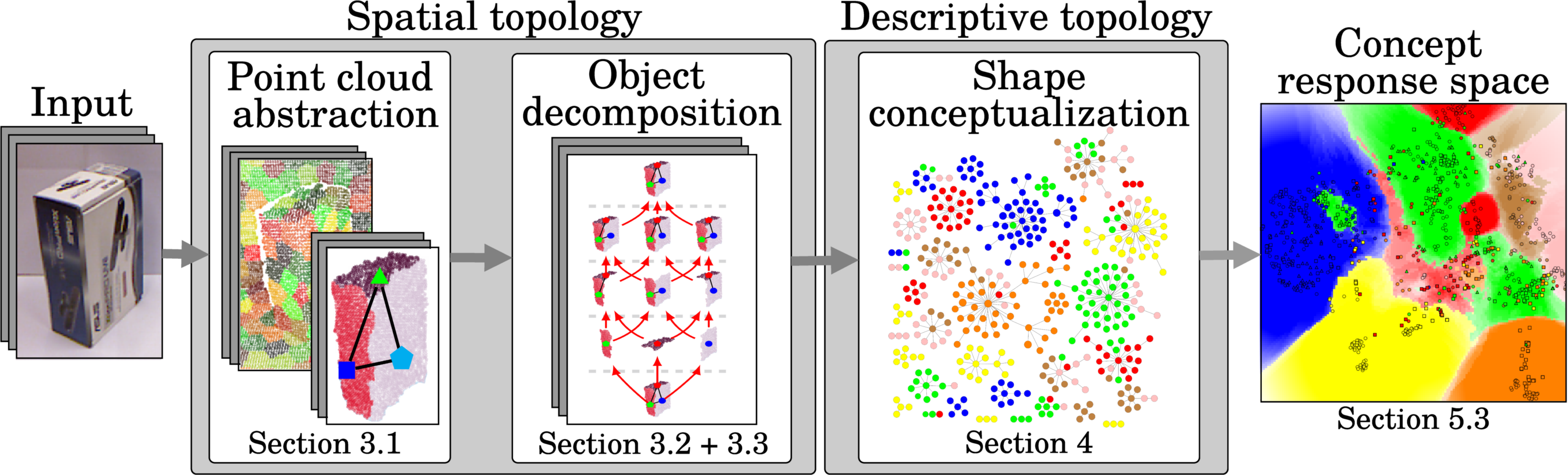}
 	\caption{Illustration of the proposed object shape conceptualization approach.}
 	\label{fig:approach_illustration}
 \end{figure}

Shape analysis relies on a robust \emph{description} and \emph{representation}~\cite{dicarlo:tics_2007} of objects, particularly in real world scenarios where captured objects are distorted by sensor noise and occlusions. 
Early cognitive and psychology theories on object perception have suggested a hierarchical and component-based representation of object information~\cite{Fodor1988}.
Inspired by this methodology, our work focuses on the analysis of topological patterns in object point cloud data observed from single viewpoints with a Kinect-like camera.
The analysis is two-fold (see Fig.~\ref{fig:approach_illustration}): \textbf{i)} analysis of the spatial topology in point cloud decompositions, \textbf{ii)} topology analysis of these decompositions in description space.
Concerning \textbf{i)}, a hierarchical decomposition of point clouds is proposed in a bottom-up manner: point clouds are initially over-segmented~\cite{Papon13CVPR} and further post-processed to segments that can reflect meaningful components of objects.
These segment compositions of objects allow to reason about shape characteristics and commonalities; commonalities observed within objects can be generalized to a shape concept.
Constellation models which learn concepts from perceived feature (e.g. keypoints or segments) constellations have shown success in recent years~\cite{6942984,local_ContextSemanticLabelling-IJRR13,7139358}.
The inference is often based on \emph{local} analysis of feature coherences with the a priori learned constellation model, i.e. local evidences in a constrained spatial range w.r.t. the features using e.g. Markov Networks~\cite{citeulike:8742196}.
Such inference is robust to absence of features due to noise and partial object occlusion.
In contrary, shape facets which become apparent on \emph{global} scale -- especially in case of complex structures -- are insufficiently reflected considering local inferences.
In this work a hierarchical constellation model is proposed in which segment constellations are decomposed over multiple topological levels that gradually (from local to global) reflect shape facets: from individual segment occurrences over segment groups to a single group of segments which represent an entire object.
On each topological level shape characteristics are observed and learned.
Concerning \textbf{ii)}, observed decompositions over these topological levels are analyzed to gather distinctive insights and patterns that can be interpreted and related to concepts of specific shape appearances.
\emph{Persistent Homology} (PH) is a concept related to Topological Data Analysis that has been applied in various areas such as related to high dimensional data visualization or finding relations and coherencies in \emph{Big Data} scenarios~\cite{carlsson2014}.
PH allows to extrapolate features from data by means of finding persistent (or stable) feature appearances through an iterative filtration of the data. 
This procedure allows to investigate the topological evolution of the data in a step-wise manner compared to standard clustering approaches such as \emph{k-Means}, \emph{Expectation-Maximization} or tree-based algorithms which are often parameterized by thresholds.
The concept of PH has shown its applicability in geometric shape analysis to detect persistent shape patterns when directly applied on point cloud data~\cite{carlsson2014,6909654}.
Instead of PH directly applying on point cloud data, we focus on responses which are retrieved from the topological analysis proposed in \textbf{i)}.
The evolutionary PH-based analysis allows to detect persistent appearances of these responses during the filtration process which reveal shape commonalities of instances that can form concepts.

\section{Spatial Topology Analysis}
\label{sec:spatial_topo_analysis}
In the following a description and representation of object segment constellations is proposed that hierarchically encodes shape facets of objects.

\subsection{Object Segment Extraction}
\label{sec:obj_seg_dict}
\begin{figure}
    \small
	\centering %
	\subfigure[Point cloud abstraction ]{\label{fig:lower_segmentation}\includegraphics[width=0.485\linewidth]{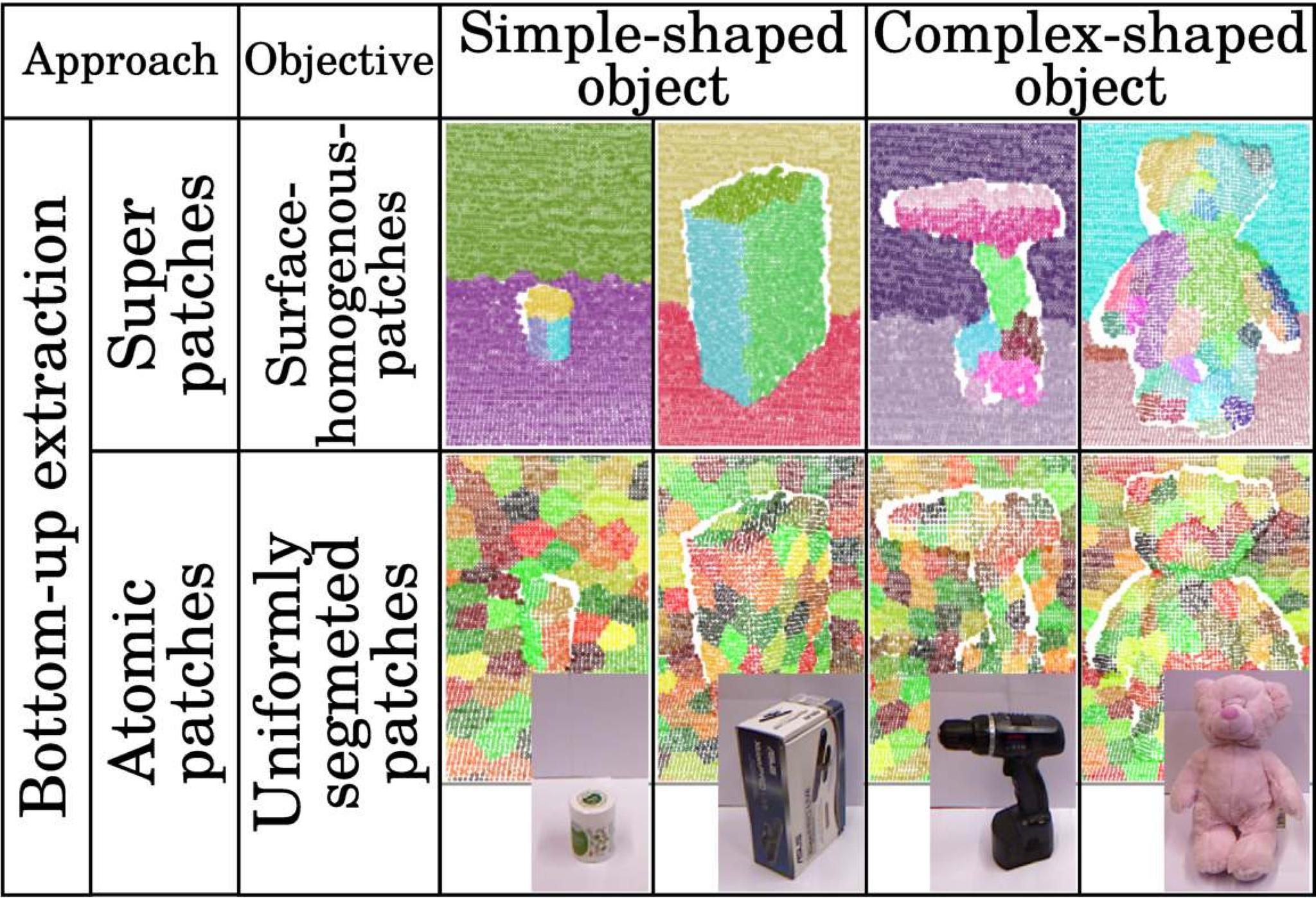}}
	\subfigure[Hierarchical dictionary]{\label{fig:dictionary}\includegraphics[width=0.5\linewidth]{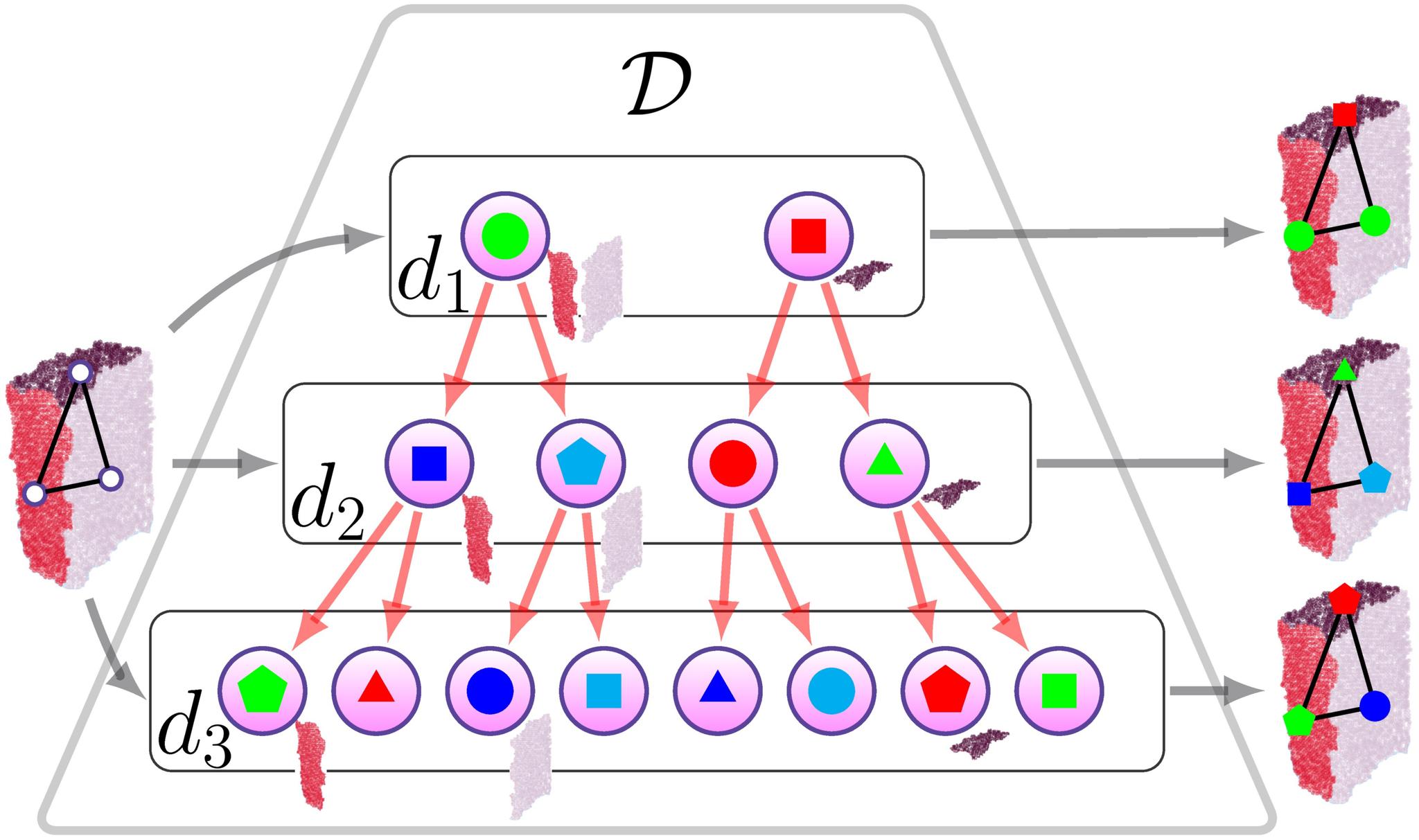}}
	\caption{In \subref{fig:lower_segmentation}, a two-step segmentation proposed in \cite{MuellerBirkIcra2016}. From atomic patch segments to super patches -- illustrated by samples (\emph{can}, \emph{box}, \emph{cordless drill}, \emph{teddy}). In \subref{fig:dictionary} an example hierarchical dictionary~\cite{6942984} $\mathcal{D}\mathrm{=}\{d_1,d_2,d_3\}$ consisting of 3 levels using divisive clustering. For illustration, each visual word is depicted as a circle with a colored polygon.} 
	\label{fig:eval:tsne_cfmat}
\end{figure}

Our presented work~\cite{MuellerBirkIcra2016} particularly focuses on noisily captured point cloud data from real world observations.
An object point cloud is initially \emph{over-segmented} (atomic patches) and further processed to segments (super patches) which can represent semantically meaningful shape components like planar surfaces of a \emph{box} or cylindric and planar surfaces of a \emph{can} (see Fig.~\ref{fig:lower_segmentation}).
Subsequently, objects are represented as a set of \emph{point cloud segments}.
These segments can be interpreted as \emph{building blocks} that constitute objects.
To mitigate the Correspondence Problem among distorted segments cause by sensor noise, a \emph{symbolic} representation of segments is chosen as an abstraction step to facilitate further shape reasoning. 
Segment appearances are quantized to a set of discrete so-called visual words following the Bag-of-Words methodology~\cite{6942984}; visual words constitute a so-called \emph{dictionary}.
The idea is that similar appearing segments are abstracted to the same symbol, respectively, visual word.
The level of quantization plays a crucial role, since too few words may lead to underfitting, whereas too many words may lead to overfitting symptoms.
For an unbiased and purely data-driven word generation, a hierarchical divisive clustering procedure is applied as proposed in our previous work~\cite{6942984}.
Therein segments are initially described with a \emph{description vector} that is generated by a point cloud descriptor like FPFH~\cite{5152473}.
As a result of the clustering procedure a hierarchical dictionary $\mathcal{D}$ is created that consists multiple description levels $\{d_1, d_2,...\}$, where level $f$ consists of $2^f$ words, see Fig.~\ref{fig:dictionary}.
Each word represents a description vector whose position is inferred by the clustering procedure during the training phase using a set of segments captured from random scenes. 

Given an object segment, the extracted description vector of the segment is passed through the hierarchical dictionary $\mathcal{D}$. 
For each description level, the propagated description vector is accordingly labeled with the visual word whose description vector is closest using $l^2$-norm; see augmented segments with words on the right side in Fig.~\ref{fig:dictionary}. 

\subsection{Hierarchical Object Decomposition and Representation}
A segment composition of object $o$ is initially represented as graph $g^o$ in which each segment represents a vertex and neighboring vertices are connected with an edge.
Each vertex is augmented with the corresponding segment point cloud and visual word that is inferred from the set of visual words of the respective description level in dictionary $\mathcal{D}$ (see Sec.~\ref{sec:obj_seg_dict}); visual word inferences differ according to the description level as shown on the right side in Fig.~\ref{fig:dictionary}.

\begin{figure}[tb]
    \small
	\centering
	\begin{minipage}[b]{.38\linewidth}
		\subfigure[Shape motif hierarchy $\mathcal{H}$]{\label{fig:ch_illustration}\includegraphics[width=1.0\linewidth]{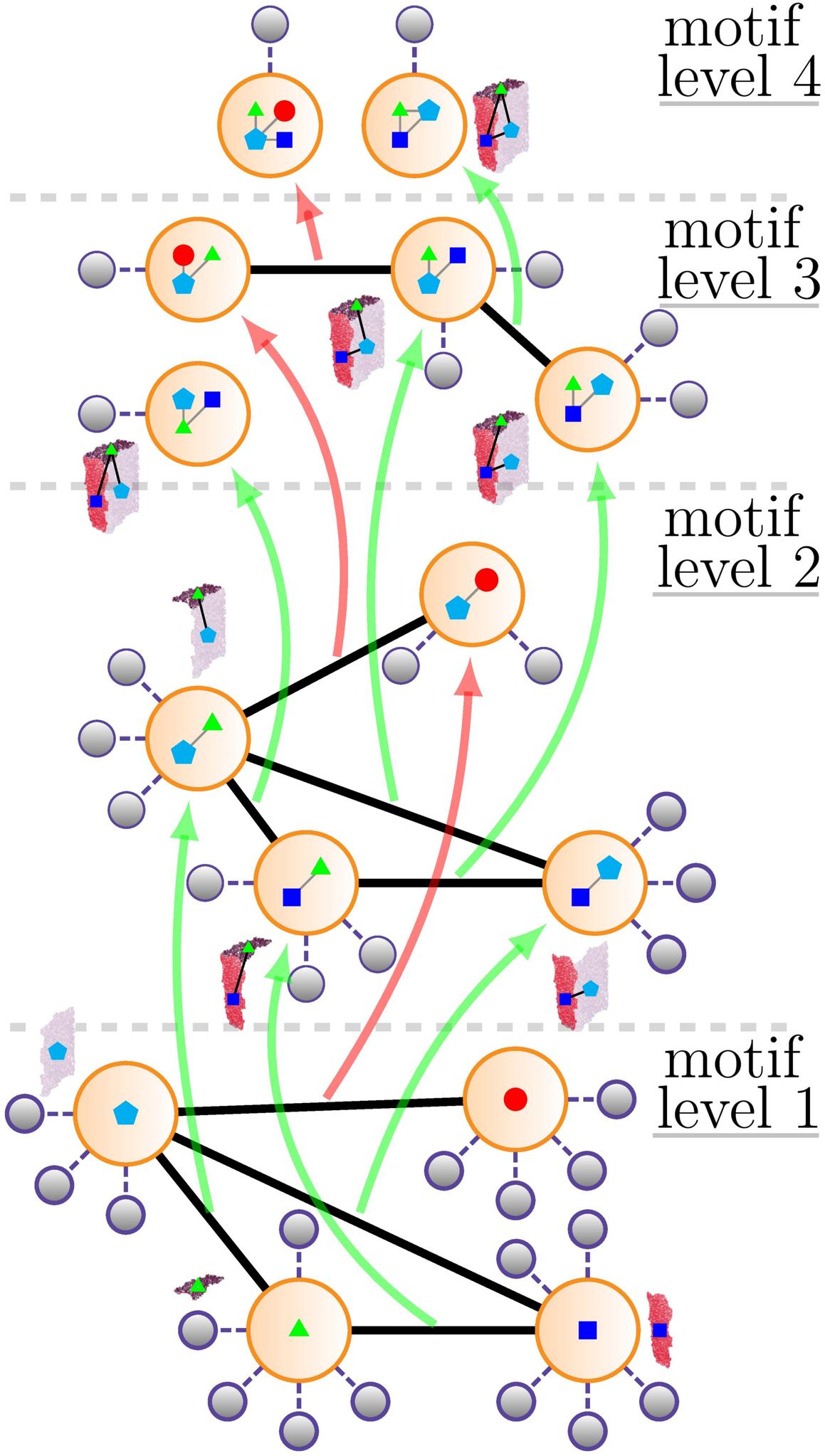}}
	\end{minipage}
	\begin{minipage}[b]{.57\linewidth}
		\centering
		\subfigure[Shape motif level]{\label{fig:ch_legend}\includegraphics[width=0.78\linewidth]{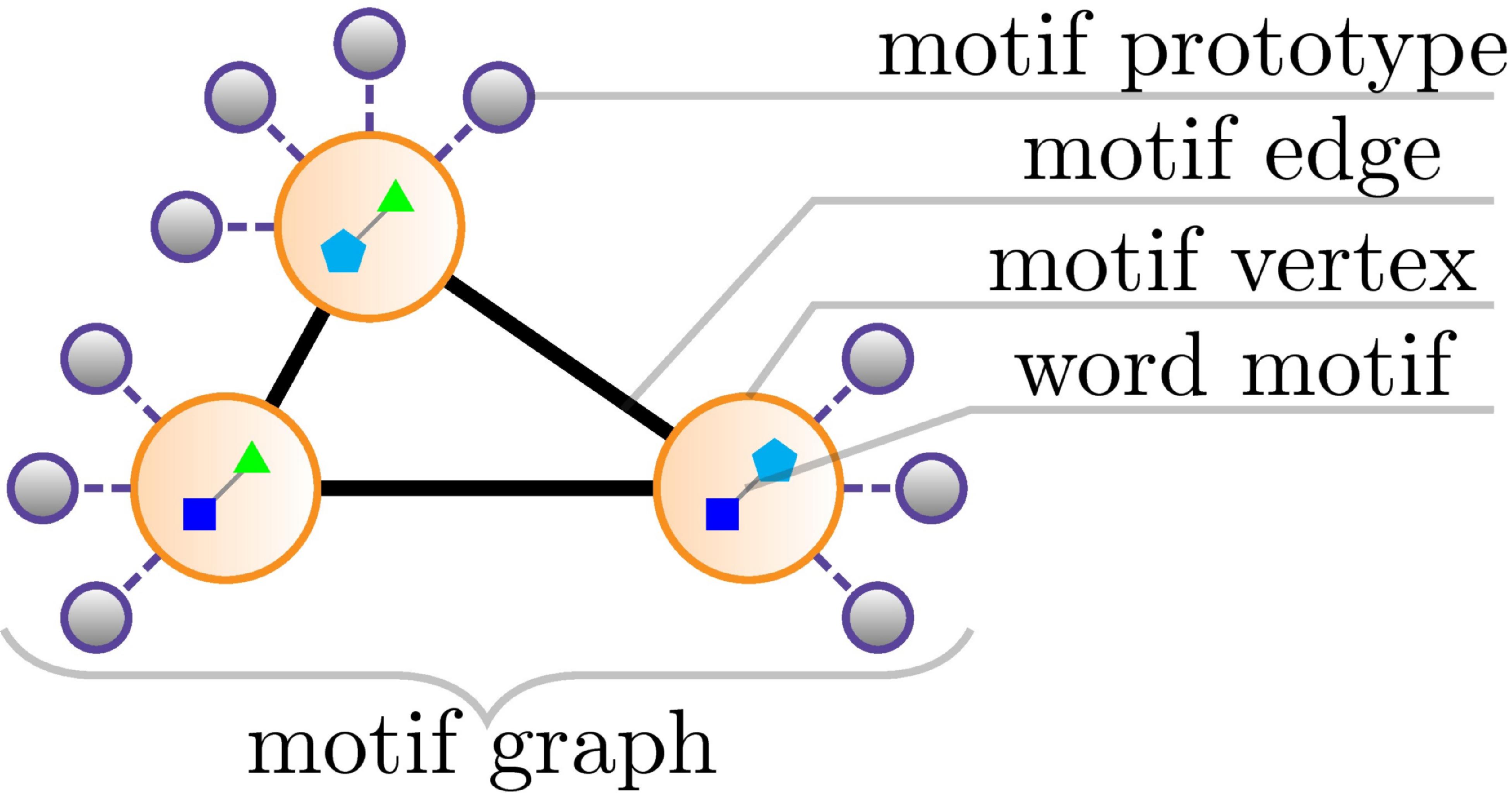}}
		
		\subfigure[Shape motif hierarchy ensemble]{\label{fig:hch_illustration}\includegraphics[width=1.0\linewidth]{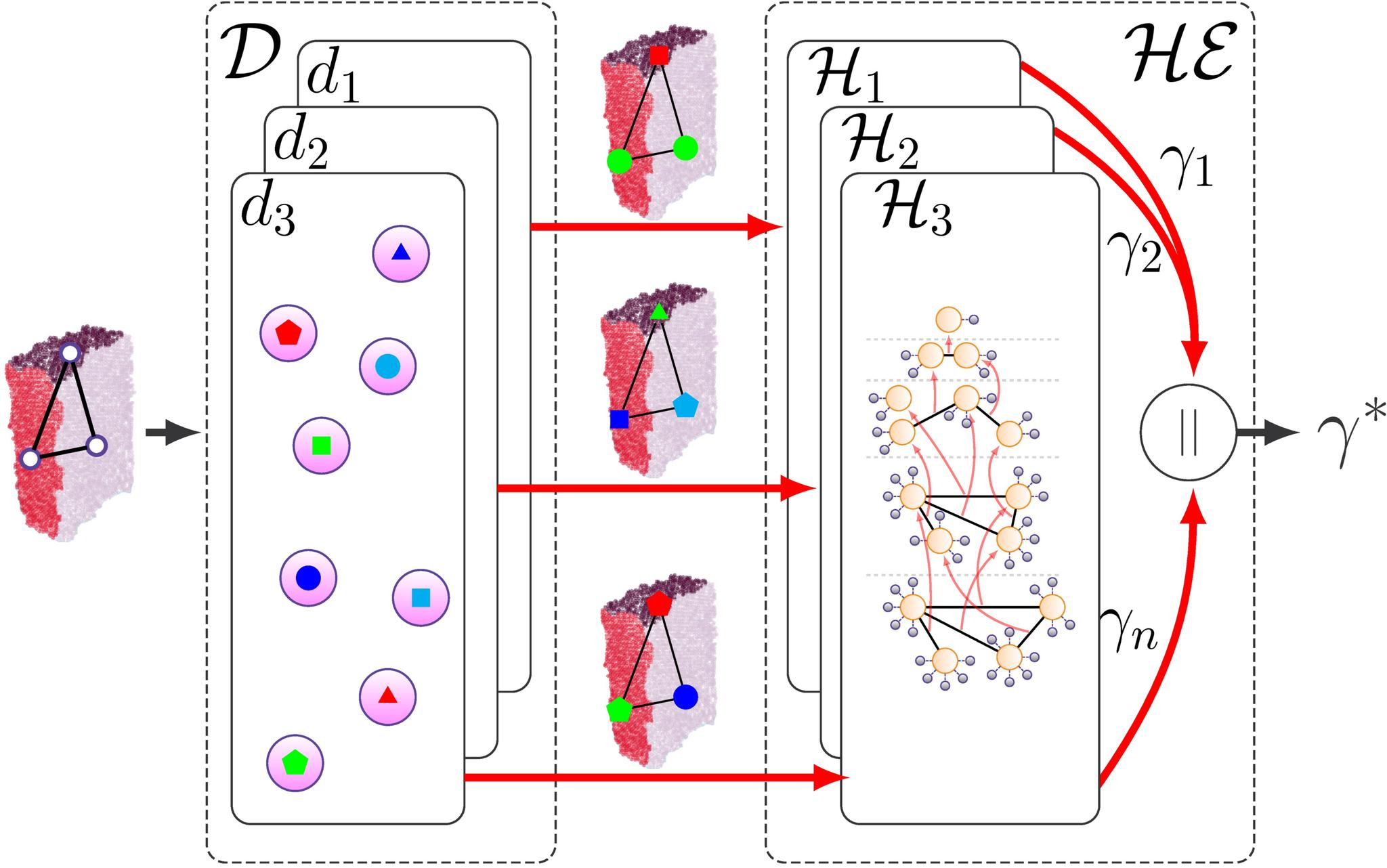}}
	\end{minipage}
	\caption{A shape motif hierarchy $\mathcal{H}$ example is shown in \subref{fig:ch_illustration}, consisting of multiple \emph{motif levels}. Each node \inlineimage{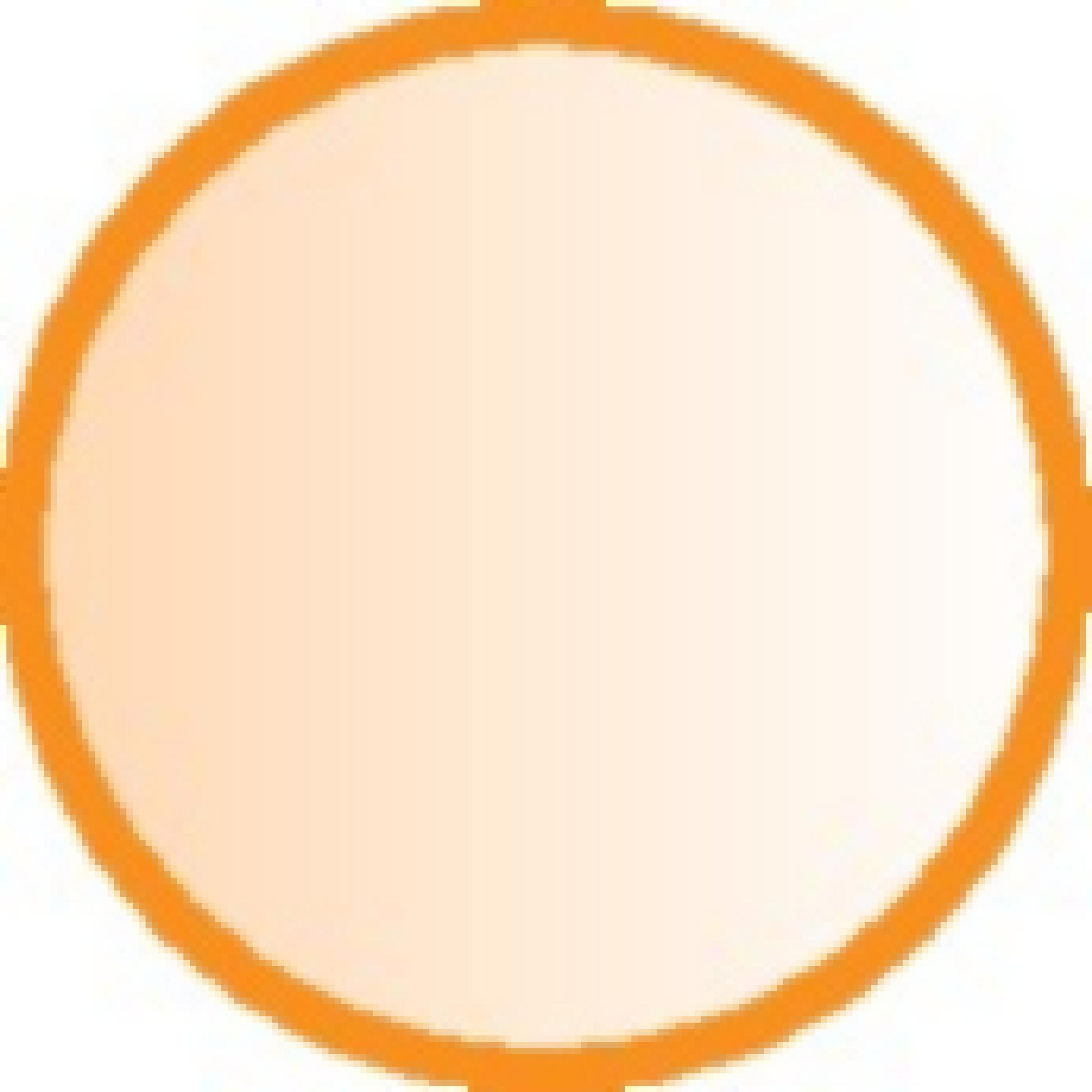} represents a specific \emph{motif vertex}, whereas each smaller linked node \protect \samplePattern\ represents a \emph{motif prototype}. A sample propagation (\protect \inlineimage{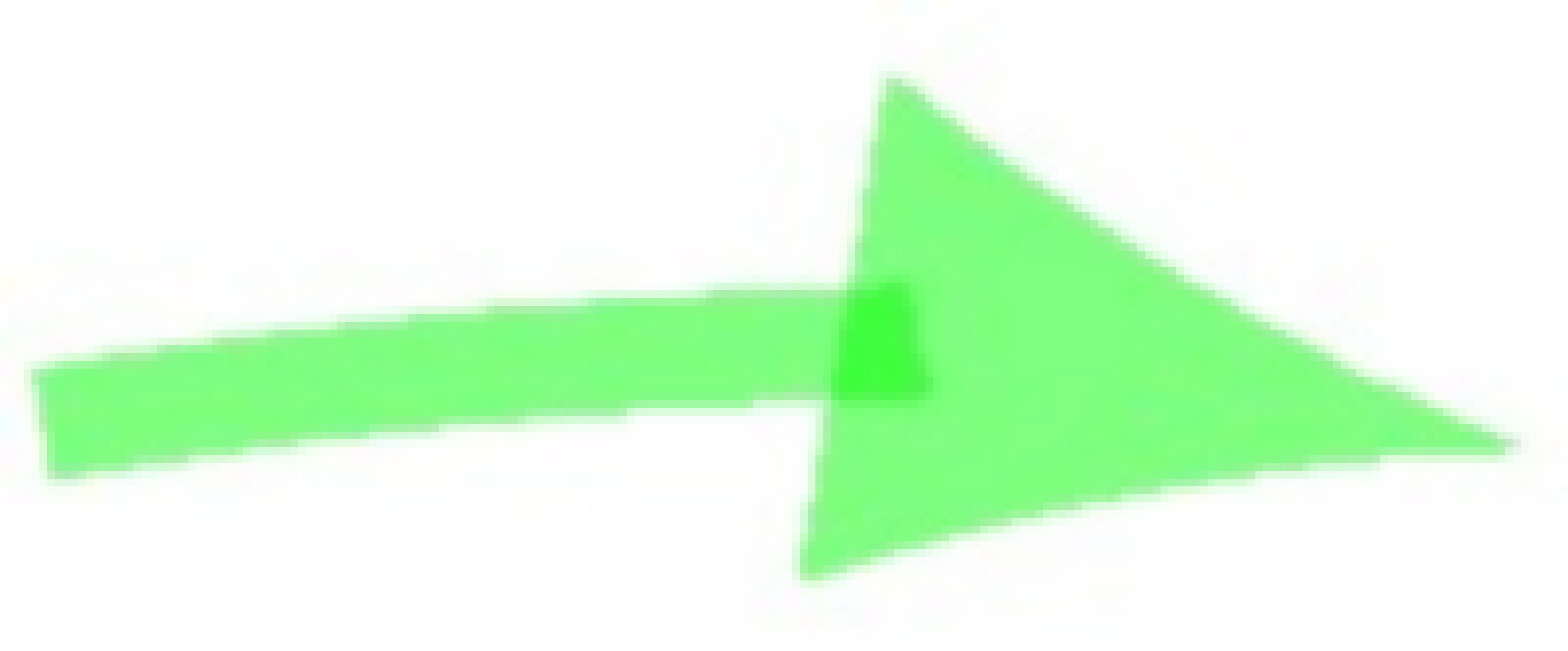}) of a box \inlineimage{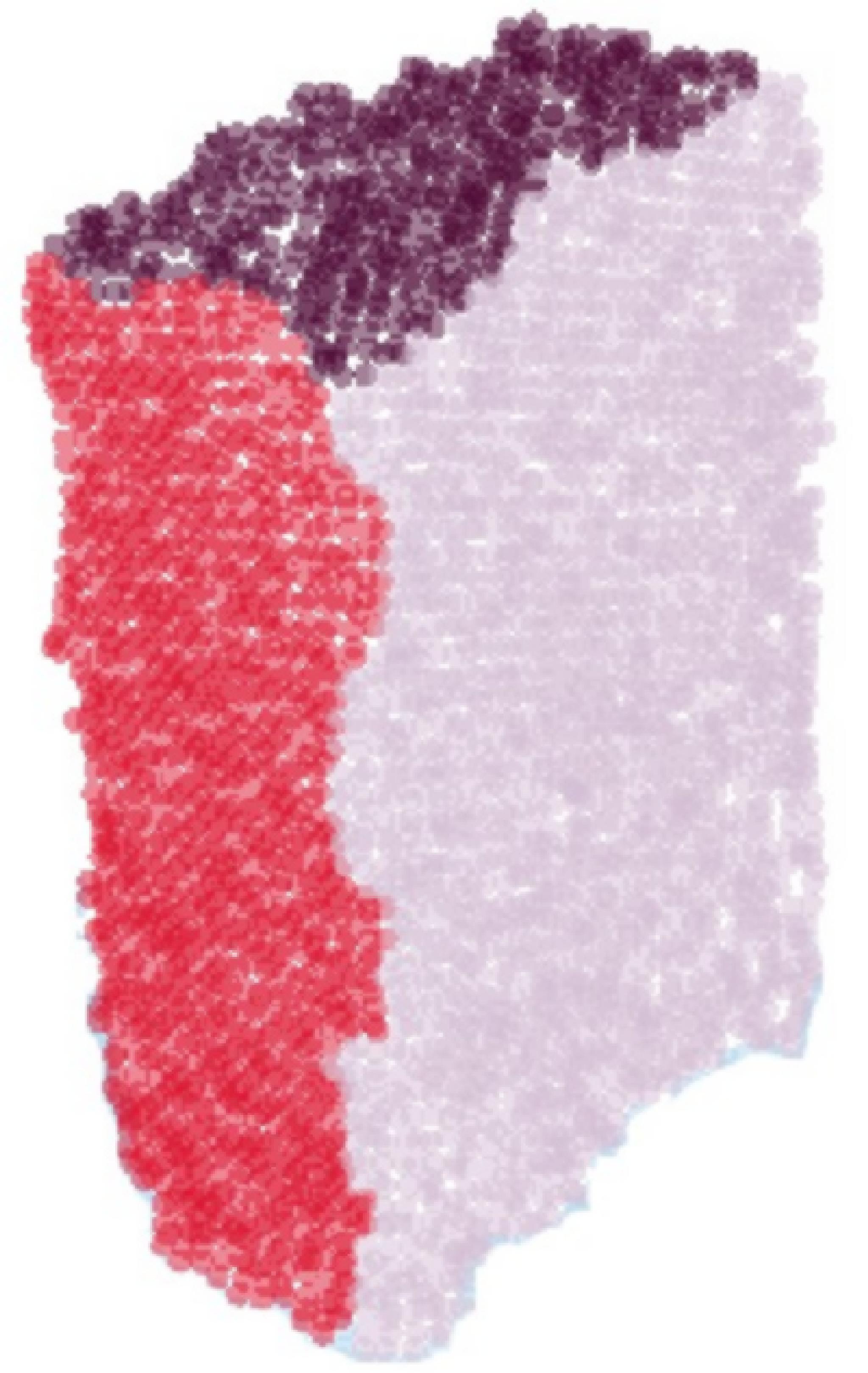} (consisting of three segments) through $\mathcal{H}$ is shown in \subref{fig:ch_illustration}.
		Feasible propagations which have been previously encoded in the hierarchy during training phase but are not affected by the \emph{box} are depicted as \protect \inlineimage{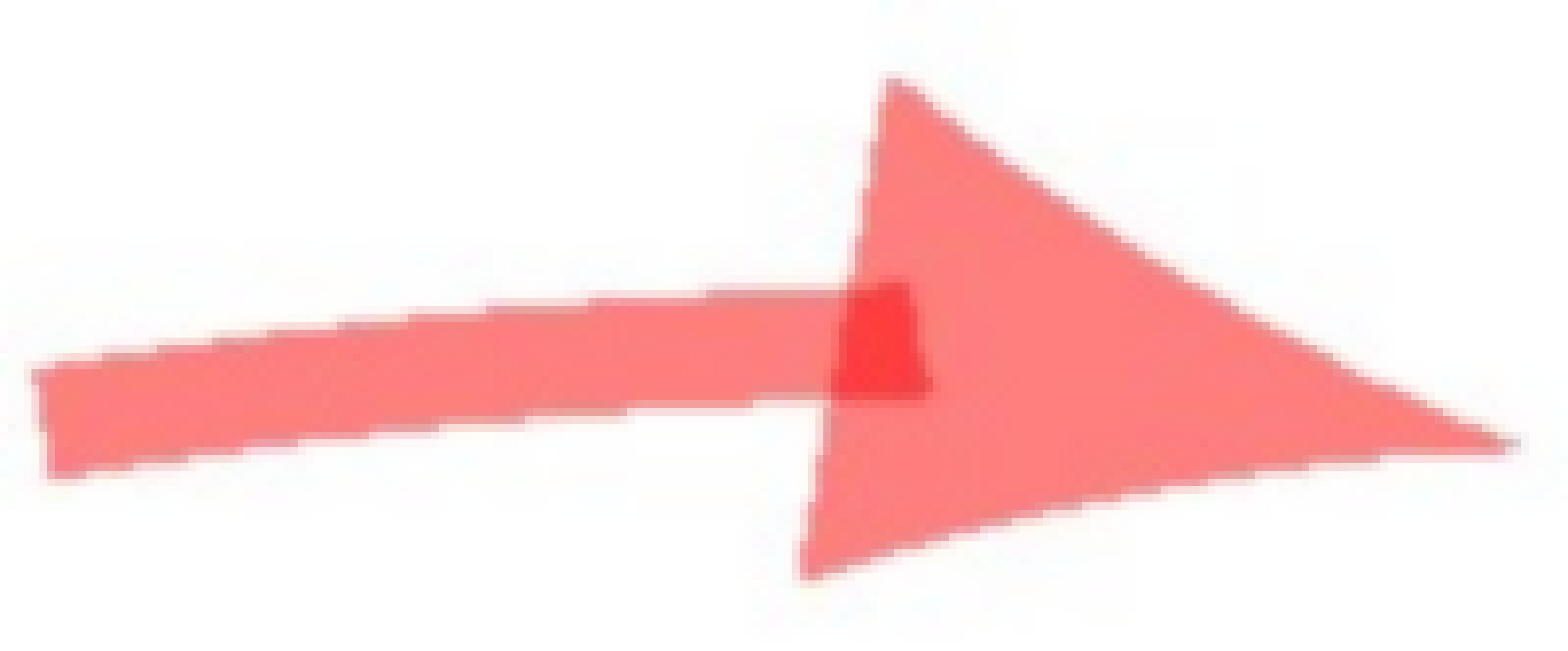}.
		Components of a \emph{motif level} are illustrated in \subref{fig:ch_legend}.  
		In \subref{fig:hch_illustration} the combined approach is illustrated:
		an example shape motif hierarchy ensemble $\mathcal{HE}$ based on three shape motif hierarchies $\{\mathcal{H}_1, \mathcal{H}_2 ,\mathcal{H}_3\}$ using respective description levels $\{d_1, d_2, d_3\}$ of $\mathcal{D}$ (see Fig.~\ref{fig:dictionary}).
	}
	\label{fig:dict_illustration_and_ch}
\end{figure}

The spatial topology of segments is unsupervisedly analyzed and encoded in a hierarchical representation which we denote as \emph{Shape Motif Hierarchy}; an illustration of a hierarchy $\mathcal{H}$ is shown in Fig.~\ref{fig:ch_illustration}.
$\mathcal{H}$ is based on a graphical representation of visual word constellations which are denoted as \emph{motifs}; note that these constellations can only contain visual words of a specific description level. 
Therefore for a dictionary $\mathcal{D}\mathrm{=}\{d_1,d_2, ..., d_n\}$ which contains $n$ description levels, $n$ hierarchies are created that constitute an ensemble  $\mathcal{HE}\mathrm{=}\{\mathcal{H}_1,\mathcal{H}_2, ..., \mathcal{H}_n\}$, see Fig.~\ref{fig:hch_illustration}.

In the training phase for each hierarchy $\mathcal{H}$, object observations are encoded in a bottom-up manner, beginning with single object segments over groups of segments until a single constellation of segments represents the entire object; a sample propagation (\protect \inlineimage{sample_ch_prop_object}) of a box \inlineimage{sample_object_a} (consisting of three segments) through the hierarchy $\mathcal{H}$ is shown in Fig~\ref{fig:ch_illustration}.
Object segments are propagated through the hierarchy considering the corresponding visual words associated to the segments.
Within the propagation process, newly observed visual word constellations (\emph{word motifs}) are integrated to the hierarchy as \emph{motif vertices} (see Fig.~\ref{fig:ch_legend}).
Each motif in the hierarchy is unique w.r.t. visual words, i.e. a newly observed word motif of an object leads to a creation of a \emph{motif vertex} if the motif does not exist in the hierarchy.
As for further characterization of a motif vertex, 
a point cloud description is extracted of a propagated segment constellation and added as \emph{motif prototype} \samplePattern\ to the motif vertex \inlineimage{sample_clique_vertex_a} that corresponds to the motif of the propagated constellation (Fig.~\ref{fig:ch_legend}).
As a result each motif vertex represents a \emph{shape motif} that can be exploited as \emph{building block} and can constitute -- even unknown -- objects.
Further on at motif level $l\mathrm{=}1$, edges (\inlineimage{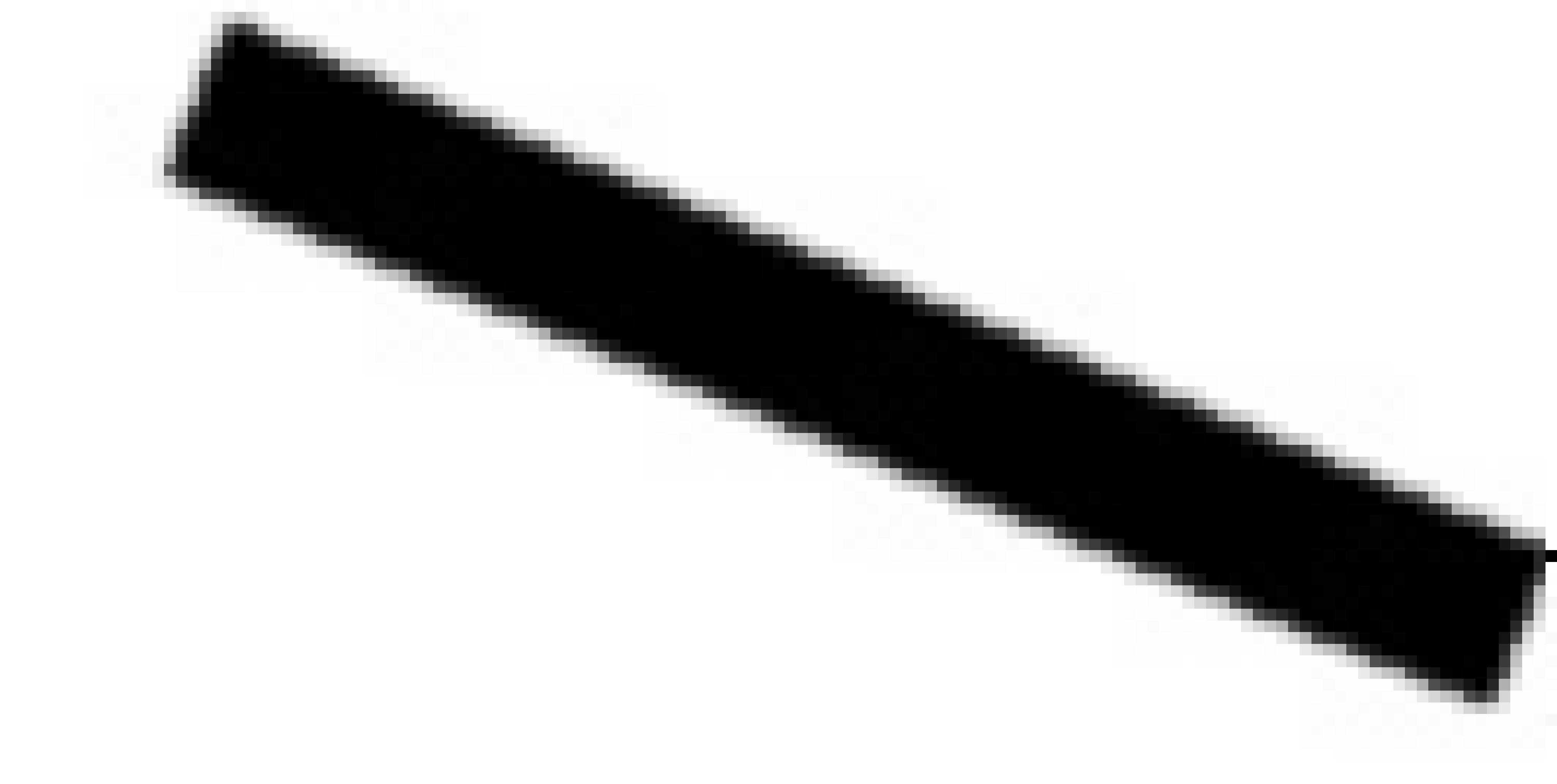}) between two motif vertices are created, if corresponding object segments are neighbors, whereas for $l\mathrm{>}1$ if two motif vertices contain a visual word that corresponds to the \emph{same} segment of the propagated object.
In each propagation step from level $l$ to $l\mathrm{+}1$, the union of word motifs connected to an edge in level $l$ form a vertex in $l\mathrm{+}1$ (\protect \inlineimage{sample_ch_prop_object}).
Consequently, upper levels can constitute of fewer edges or vertices, i.e. a single motif vertex can encompass a word constellation that represents an entire object, e.g. see \emph{box} sample \inlineimage{sample_object_a} at motif level 4 in Fig.~\ref{fig:ch_illustration}.
In this manner, objects are decomposed in various motifs by the propagation through hierarchy $\mathcal{H}$.

As a result, the Shape Motif Hierarchy Ensemble $\mathcal{HE}\mathrm{=}\{\mathcal{H}_1,\mathcal{H}_2, ..., \mathcal{H}_n\}$ does not only consider the structural appearance regarding segment constellation variety but also the symbolic appearance of constellations by using a particular dictionary description level for the respective hierarchy.

\subsection{Stimuli Generation} %
\label{sec:stimuli_generation}
In the training phase object segment constellations associated with corresponding visual words are propagated through the hierarchy and are memorized as \emph{motif prototypes} within motif vertices that matches visual word constellations of the object. %
Inspired by the \emph{Prototype Theory}~\cite{Rosch1973}, each motif vertex is constituted of these prototypes which are exploited to generate stimuli for unknown objects as described in the following: given a graph of segments $g^o$ of object $o$, these segments are annotated with the corresponding words and subsequently propagated through the hierarchy as in the training phase, see \emph{box} in Fig.~\ref{fig:ch_illustration}; note that the hierarchy is not modified while stimuli generation.
Through the propagation of segments, motif vertices are activated which correspond to the words of the propagated segments.
An \emph{activation} of a vertex $v$ is represented by the Indicator function $\mathds{1}_v(g^o)$ which returns $1$ in case of a match, otherwise $0$ if no match is found.
For an activated $v$ a stimulus $\alpha(v,g^o)$ is computed based on point cloud descriptions of the memorized \emph{motif prototypes} $T^v$ of $v$ and the respective description $q$ of object segments in $g^o$
which activated $v$.
By applying the approach of Probabilistic Neural Networks~\cite{Huang:2004:APN:1011980.1011984}, the stimulus is computed with an adapted Gaussian kernel (e.g., bandwidth $\sigma\mathrm{=}0.025$) in which Jenson-Shannon divergence~($JSD$) is used as distance measure, see Eq.~\ref{eq:stimuli}.
\begin{equation} \label{eq:stimuli}
\small
\alpha(v,g^o) = 
\begin{cases}
    \frac{1}{|T^v|} \cdot \sum^{|T^v|}_{i=1}e^{\tfrac{\mathrm{JSD}(t_i \in T^v,q)^2}{-2\sigma^2}},&\text{\footnotesize if $\mathds{1}_v(g^o)\mathrm{=}1$}\\
    0,              & \text{\footnotesize otherwise}
\end{cases}
\end{equation}
As a result for each propagated object, stimuli of motif vertices in $\mathcal{H}_i$ are accumulated and projected into vector form $\gamma^o_i\mathrm{=}[\alpha(v_1,g^o),  \alpha(v_2,g^o), ...]$.
Given $n$ description levels and correspondingly trained $n$ shape motif hierarchies that form the ensemble $\mathcal{HE}\mathrm{=}\{\mathcal{H}_1,\mathcal{H}_2, ..., \mathcal{H}_n\}$, the object graph $g^o$ is propagated through each motif hierarchy. 
Subsequently, a final stimuli vector $\leftidx{^*}\gamma^o\mathrm{=}[\gamma^o_1$, $\gamma^o_2$, $..., \gamma^o_n]$ is composed ($||$) of stimuli retrieved from $n$ motif hierarchies, see Fig.~\ref{fig:hch_illustration}.

\section{Descriptive Topology Analysis}
\label{sec:desc_topo_analysis}
Commonalities among shape appearances can vary from \emph{specific} to \emph{generic} shape facets: a concept generation process is proposed that in a gradual manner reveals commonalities ranging from individual to common facets.
Persistent Homology (PH) provides a computational model that allows to gradually reveal topologically persistent patterns in  generated stimuli $\leftidx{^*}\gamma$ which can be interpreted as commonalities and eventually as shape concepts.

\subsection{Persistence Homology and Filtration}
\label{sec:filt_PH}
We briefly introduce terms from algebraic topology for our major goal of extracting shape concepts.
Comprehensive literature can be found in \cite{carlsson2014,Edelsbrunner2002}.

\subsubsection{Simplices and Complexes}
Given a continuous topological space $\mathcal{X}\mathrm{=}\{ x_0, x_1, ..., x_m| x_i\in \mathcal{R}^n, 0 <= i <= m\} $ with $n$-dimensional $m$ data points.
A \emph{simplex} $\pi$ is a $d$-dimensional polytope which is a graph consisting of a convex hull of $d\mathrm{+}1$ affine independent vertices where each vertex is a point in $\mathcal{X}$.
A composition of \emph{simplices} is denoted as \emph{simplicial complex} $K\mathrm{=}\{\pi_0, \pi_1, \pi_2, ...\}$.
This composition is a union of vertices, edges, triangles or other higher dimensional polytopes.
\subsubsection{Vietoris-Rips Complex}
We focus on \emph{vietoris-rips complexes} in which a complex $K^{vr}_i$ is extracted from a subspace $\mathcal{X}_i \mathrm{\subseteq} \mathcal{X}$ with a given scale parameter $\epsilon \mathrm{>} 0$.
$K^{vr}_i$ consists of vertices which are only connected if the distances between the vertices is lower than the given parameter $\epsilon$.
The vietoris-rips complex $K^{vr}_i$ can also be denoted as $\epsilon$\emph{-complex}, where $\epsilon$ is also denoted as radius or distance threshold.

\subsubsection{Homology Groups}
Homology is a concept in \emph{algebraic topology} which allows to reveal specific characteristics or features in $\mathcal{X}$.
Therein characteristics are organized into so-called homology groups $\mathcal{HG}\mathrm{=}\{H_0(\mathcal{X}), H_1(\mathcal{X})$, $H_2(\mathcal{X}), ...\}$.
Often the first three homology groups are analyzed: in the context of geometry $H_0(\mathcal{X})$ is related to \emph{connected components} or \emph{clusters} of vertices, $H_1(\mathcal{X})$ is related to the complexes in form of \emph{loops} or \emph{holes} and $H_2(\mathcal{X})$ is related to so-called \emph{voids} which represent fully connected complexes.
In this work we focus on $H_0(\mathcal{X})$ since it complies with our goal to extract topological groups from stimuli vectors (see Sec.~\ref{sec:stimuli_generation}) which can represent concepts.
\subsubsection{Topological Space Filtration and Persistent Homology}
\label{sec:topo_filration}
The filtration of the topological space $\mathcal{X}$ is initiated by a subsequently nested application of a set of radii $\mathcal{E}\mathrm{=}\{\epsilon_0, \epsilon_1, ..., \epsilon_j\}$ where $\epsilon_{i-1} \mathrm{<} \epsilon_i \mathrm{<} \epsilon_{i+1}$.
In case of $H_0(\mathcal{X})$, at the beginning of the filtration process each point $x_i \in \mathcal{X}$ is represented by a 0-simplex $\pi_i \in $ \emph{vietoris-rips complexes} $K^{vr}_0$.
These created simplices are \emph{born} at radius $0$.
Note that the $K^{vr}_i$ is extracted using radius $\epsilon_i$.
While the filtration progresses, the vietoris-rips complex grows since the radius increases which can cause fusions of simplices that form a larger simplex: a \emph{union} is performed between simplices while one simplex enlarges and sustains by annexing the other that \emph{dies}.
Eventually, a complex $K^{vr}$ is filtered that contains a single high dimensional simplex -- see Eq.~\ref{eq:filtration}.
\begin{equation}
\small
\emptyset \subseteq K^{vr}_0 \subseteq K^{vr}_1 \subseteq \dotso \subseteq K^{vr}_j = K^{vr}
\label{eq:filtration}
\end{equation}
Persistent Homology is a way to analyze and track birth and death of simplices (a.k.a. \emph{homology classes}) along the filtration process: $H_0(K^{vr}_i) \rightarrow  H_0(K^{vr}_{i+1})$.
Results are represented in \emph{persistence} or \emph{barcode diagrams} (see Fig.~\ref{fig:barcode}).
While considering the gradual evolution of vietoris-complex $K^{vr}_i$, the extraction of homology classes (birth and death) is inherently robust to deformation due to the topological organization of the data in a graphical manner.

\subsection{Shape Concept Extraction}
\label{sec:concept_extraction}

\subsubsection{Topological Space Generation}
\label{sec:topo_space_gen}
Given a set of raw stimuli vector responses (see Sec.~\ref{sec:stimuli_generation}), the responses are initially used to create a \emph{topological space} in a graphical manner.
Therein, a stimuli vector  $\leftidx{^*}\gamma$ can be interpreted as an independent point in the space in which a distance metric can be applied to measure the similarity to other stimuli vectors; these vectors serve as anchor points in a space of unknown topology.
The goal is to interrelate these vectors in order to discover topological relationships among these anchor points.
We make use of a graphical representation, in which each anchor point represents a vertex.
Initially a \emph{complete graph} is created, where each edge between vertices is augmented with the corresponding distance; distances are inferred by the Jenson Shannon divergence~(JSD). %

\begin{figure}[tb]
   \small
  \centering
  \includegraphics[width=0.85\linewidth]{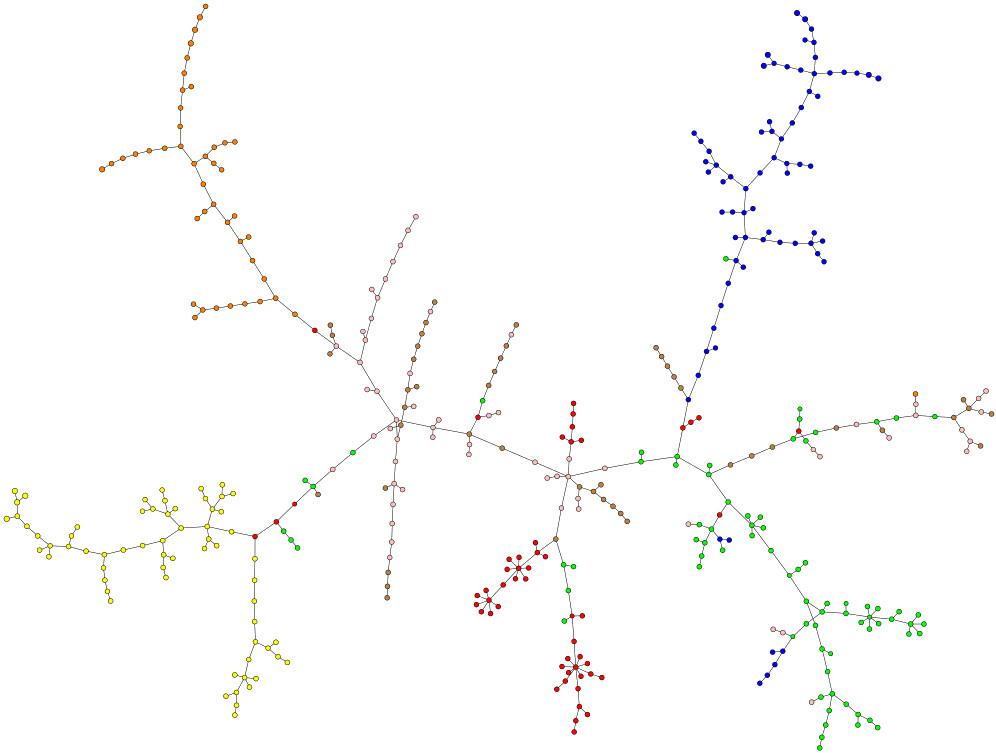}
  \includegraphics[trim=0.3cm 0.1cm 0.1cm 0.1cm,width=0.1385\linewidth]{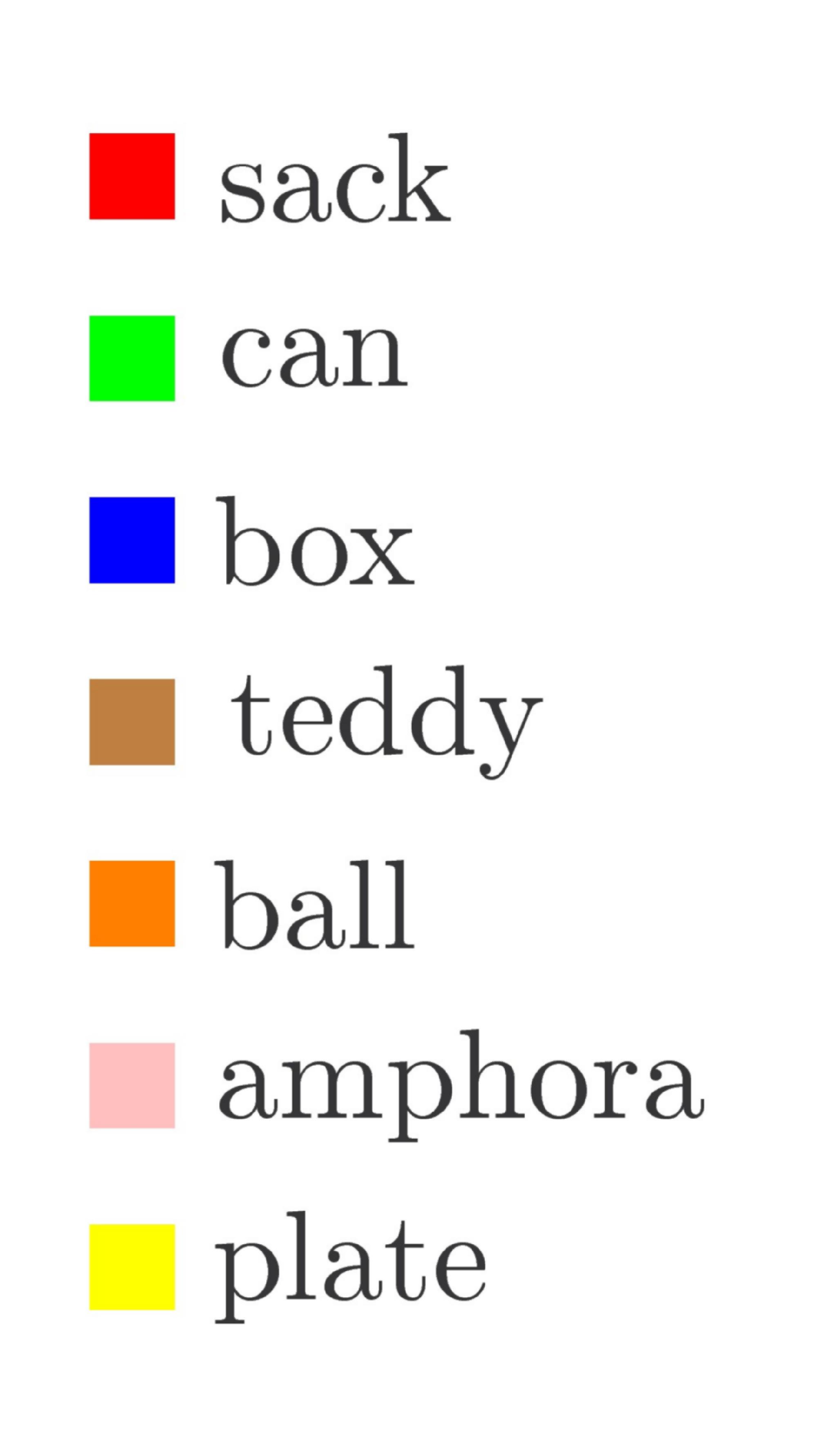}
  \caption{Minimum spanning tree which spans a topological space $\mathcal{X}$ of stimuli vectors, extracted from instances of the \emph{Object Shape Category Dataset} (see Sec.~\ref{sec:experiment} for further dataset description). Note that, each vertex represents a sample object of the dataset. Only for illustration purposes vertices are colored by their corresponding category label of the dataset.}
  \label{fig:mst}
 \end{figure}

To minimize the search space and to initiate the construction of the topological space $\mathcal{X}$, the Minimum Spanning Tree is extracted by considering the respective JSD distances.
Subsequently, a substantial amount of edges perishes and a minimum number of edges remains that allow to gain a first insight of the structural and topological organization of the stimuli vectors.
In Fig.~\ref{fig:mst}, a result is shown of object instances from the \emph{Object Shape Category Dataset} (see Sec.~\ref{sec:experiment}) that consists of seven shape categories  (\emph{sack}, \emph{can}, \emph{box}, \emph{teddy}, \emph{ball}, \emph{amphora}, \emph{plate}). 

From this point on, we focus on the \emph{topological similarity} among stimuli in form of the \emph{geodesic} distance within $\mathcal{X}$. %
Therefore each edge is uniformly weighted by assigning a distance of $1$. 
Due to the inherent sparsity of edges in $\mathcal{X}$, Johnsons all-pair-shortest path algorithm allows to efficiently generate a distance map which is used to infer a heat for each vertex $x \in \mathcal{X}$.
A vertex heat $h^{\simpcir}(x)$ is inferred by the mean geodesic distances $d_{geo}(\cdot)$ to all other vertices in $\mathcal{X}$ whereas the edge heat $h^{\ue}(e_{j,k})$ is inferred by the mean heat of the connected vertices $x_j$ and $x_k$ as shown in Eq.~\ref{eq:vertex_heat}.

\begin{equation}
\small
h^{\simpcir}(x)\mathrm{=}\frac{\sum_{i=0}^{|\mathcal{X}|} d_{\text{geo}}(x, x_i\in\mathcal{X})}{|\mathcal{X}|},\ \ h^{\ue}(e_{j,k})\mathrm{=}\frac{h^{\simpcir}(x_j)\text{+}h^{\simpcir}(x_k)}{2}
\label{eq:vertex_heat}
\end{equation}
Henceforth, we exploit edge heats as edge distances between respective vertices. 
By scaling the heat in $\mathcal{X}$ to the interval $[0,1]$ and inverting the heat,
vertices located at leaf regions of $\mathcal{X}$ become closer whereas vertices in the inner region become farther away.
Furthermore, two observations can be made: i) the heat of exteriorly located edges is lower than the interiorly located ones; ii) vertices which are interiorly located reflect more heterogeneity w.r.t their neighbors, compared to vertices which are exteriorly located in $\mathcal{X}$.

\subsubsection{Topological Filtration}
\label{sec:shape_filtration}
Given the topological space $\mathcal{X}$, the filtration is applied over a range of radii $\mathcal{E}\mathrm{=}\{\epsilon_0, \epsilon_1, ..., \epsilon_j\}$.
The step size $\epsilon_i \mathrm{\rightarrow} \epsilon_{i+1}$ is determined by the minimum edge distance in $\mathcal{X}$ which also initialize the filtration at $\epsilon_0$.
The filtration is completed when the maximum edge distance in $\mathcal{X}$ is reached at $\epsilon_j$. 
In practice the number of steps $|\mathcal{E}|$ can reach a computationally intractable number.
An upper bound limit for $|\mathcal{E}|$ can be applied by increasing the step size until the upper bound is met.
Consequently, the filtration is initialized with 0-simplices where each simplex represents a stimuli vector, and respectively, a vertex of the topological space $\mathcal{X}$.
This filtration is performed on $\mathcal{X}$ as described in Sec.~\ref{sec:topo_filration}; note that, the equidistant filtration steps from $\epsilon_0$ to $\epsilon_j$ are often denoted as \emph{time}.

\begin{figure}
   \small
   \centering
  \includegraphics[width=0.84\linewidth]{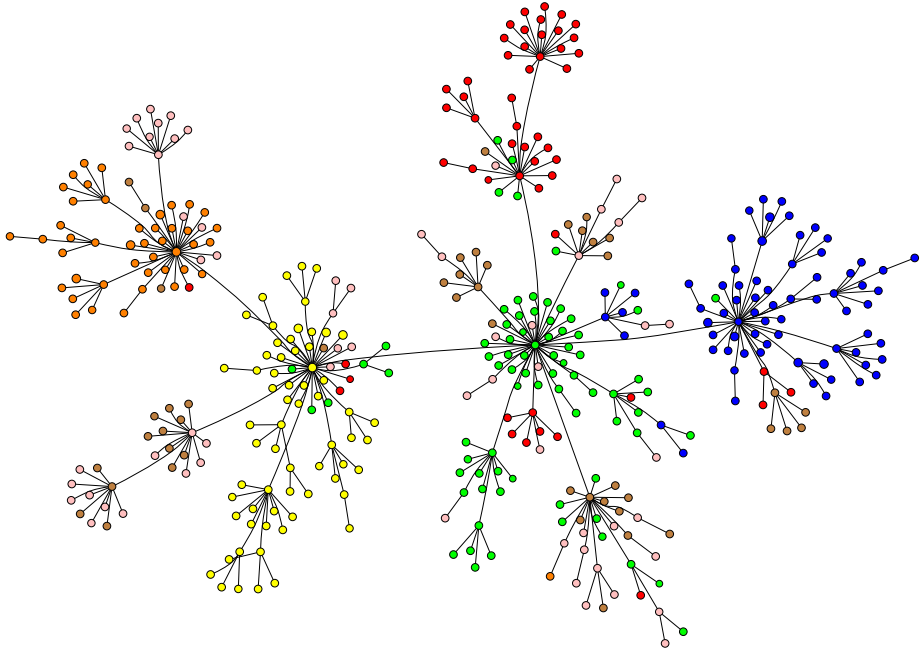}
  \caption{Filtration graph $\mathcal{F}$ showing annexations over time according to the given graph $\mathcal{X}$ (see Fig.~\ref{fig:mst}). For illustrations purposes, each vertex is colored with the corresponding label as shown in Fig.~\ref{fig:mst}}
  \label{fig:association_graph}
 \end{figure}

Persistent Homology allows to track the birth and death of simplices in $K^{vr}$ of $\mathcal{X}$ during the filtration.
Due to the nature of evolving simplices complex (see Eq.~\ref{eq:filtration}) in each time step, the complex changes its appearance after \emph{annexations} of simplices complexes of previous time steps.
These changes during the filtration are encoded in graph $\mathcal{F}$ which is shown in Fig.~\ref{fig:association_graph}.
An edge represents an annexation during the filtration process of a simplices complex to another complex -- beginning with 0-simplices representing leaves in $\mathcal{F}$. 
Each edge is augmented with the annexation time.
It can be interpreted that outer simplices have lived shorter since they have been annexed earlier in time compared to inner ones.
	As a result, $\mathcal{F}$ represents the filtration progression of $\mathcal{X}$.
\subsubsection{Persistent Shape Concept Extraction}
\label{sec:shape_concept_extraction}
The lifetime of simplices can be interpreted as feature indicator in $\mathcal{X}$, i.e. \emph{persistent} or long living simplices can represent a significant feature.
In contrary, short living simplices can be interpreted as insignificant.
The concrete goal is to detect such persistent simplices.
In order to ease the persistence analysis, the filtration time range is scaled within the interval $[0,1]$, i.e. from $0$ (start of filtration $\mathrm{=} \ \epsilon_0$) to $1$ (end of filtration $\mathrm{=} \ \epsilon_j$).
In the filtration process trivial homology classes are obtained at time $0$ where 0-simplices exist and at time $1$ where a single simplex consists of all simplices in $\mathcal{X}$.
We are interested of finding persistent groups within these extrema.
A group is a \emph{connected component} of vertices, i.e. $d$-simplex ($d\mathrm{>}0$).
Due the gradual filtration, each group consists of topological similar vertices. 
Therefore these groups can constitute shape concepts, where each vertex within a group is a representative \emph{concept prototype}.

\begin{figure}
	\small
	\centering
	\def \comImgHeight {0.25}
	\begin{minipage}[b]{.45\linewidth}
		\subfigure[]{\label{fig:ph_concept_extraction}\includegraphics[width=1.0\linewidth]{concept_association_graph_no_label}} 
	\end{minipage}
	\begin{minipage}[b]{.45\linewidth}
		\subfigure[]{\label{fig:trainsack}\includegraphics[height=\comImgHeight\textwidth]{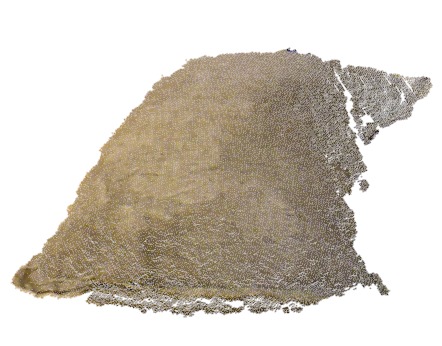}}
		\subfigure[]{\label{fig:trainbarrel}\includegraphics[height=\comImgHeight\textwidth]{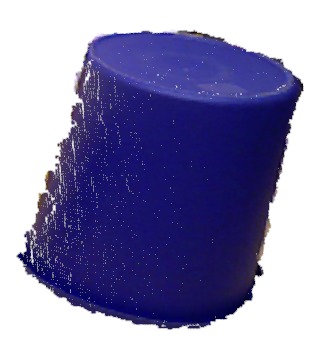}}
		\subfigure[]{\label{fig:trainparcel}\includegraphics[height=\comImgHeight\textwidth]{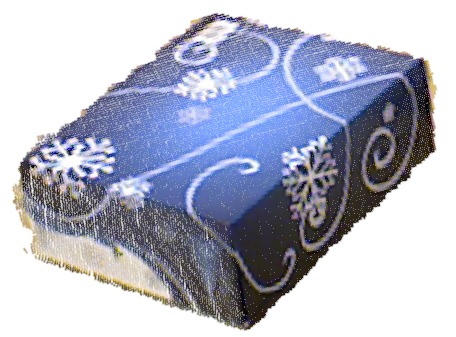}}
		\subfigure[]{\label{fig:trainteddy}\includegraphics[height=\comImgHeight\textwidth]{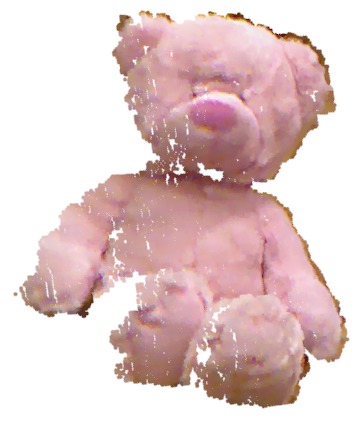}}
		\subfigure[]{\label{fig:trainball}\includegraphics[height=\comImgHeight\textwidth]{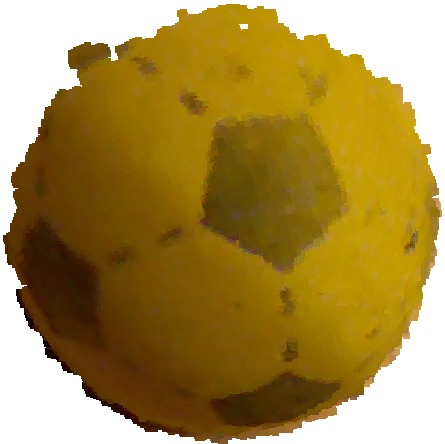}}
		\subfigure[]{\label{fig:trainamphora}\includegraphics[height=\comImgHeight\textwidth]{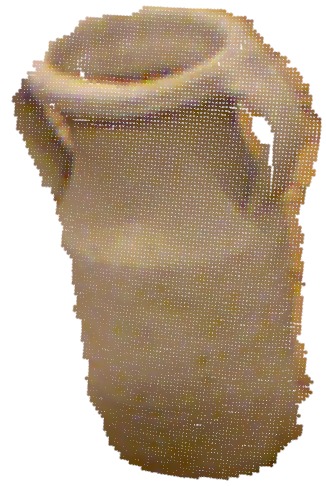}}
		\subfigure[]{\label{fig:trainplate}\includegraphics[height=\comImgHeight\textwidth]{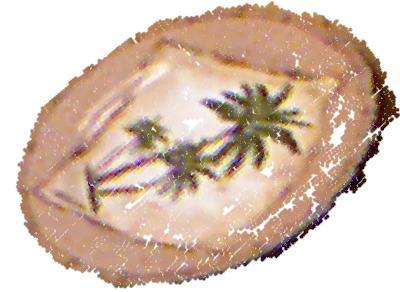}}
	\end{minipage} 
	\caption{Connected components in \subref{fig:ph_concept_extraction} extracted from $\mathcal{F}$ (see Fig.\ref{fig:association_graph}) that represent concepts $\mathcal{C}$ ($|\mathcal{C}|\mathrm{=}36$). For illustration purposes, each vertex (concept prototype) is colored with the corresponding label as in Fig.~\ref{fig:mst}. In \subref{fig:trainsack}-\subref{fig:trainplate} 2.5D point cloud sample of a random object of the OSCD dataset: \emph{sack}~\subref{fig:trainsack}, \emph{can}~\subref{fig:trainbarrel}, \emph{box}~\subref{fig:trainparcel}, \emph{teddy}~\subref{fig:trainteddy}, \emph{ball}~\subref{fig:trainball}, \emph{amphora}~\subref{fig:trainamphora} and \emph{plate}~\subref{fig:trainplate}.}
	\label{fig:ph_concept_extraction_samples}
\end{figure}

Given the entire time spectrum $[0,1]$, Persistent Homology allows to access any state of detected concepts $\mathcal{C}$ in $\mathcal{X}$ at an arbitrary time in the spectrum; note that as previously implied: the filtration starts with $|\mathcal{C}|\mathrm{=}|\mathcal{X}|$ and ends with $|\mathcal{C}|\mathrm{=}1$.
Consequently, a distinctive time can be determined.
An optimal time varies according to the topology that is reflected by the given stimuli vector and eventually by application scenario-dependent objectives.
Considering an optimal time when the global maximum of annexations (see Sec.~\ref{sec:eval:filtration}) is reached, and subsequently removing edges in $\mathcal{F}$ that are augmented with an older time than the optimal time, %
results in a set of connected components in $\mathcal{F}$ that can reflect reasonable shape concepts as illustrated in Fig.~\ref{fig:ph_concept_extraction}.
Note that, edges which are created at later time connect more heterogeneous groups and subsequently represent more generic concepts, in contrast to more specific concepts which emerge when edges are created at earlier time. 

\subsection{Shape Concept Inference}
\label{sec:shape_concept_inference}
Given a stimuli vector $\leftidx{^*}\gamma^o$ that is extracted from an unknown object $o$,  a response is retrieved based on similarity to previously learned shape concepts (see Fig.~\ref{fig:ph_concept_extraction}).
Each concept $c\in \mathcal{C}$ consists of a set of \emph{concept prototypes} $P^c\mathrm{=}\{p_1,$ $p_2, ...\}$ which are utilized to derive the correspondence of unknown objects to respective concepts.
We interpret the \emph{Prototype Theory}~\cite{Rosch1973} such that unknown instances are classified based on the similarity to known instances which are associated to the previously learned shape concepts.
To demonstrate and emphasize the discrimination capability of the proposed shape representation, the similarity $\phi^c(\cdot)$ to a concept $c$ is inferred by a (basic) mean similarity among  $\leftidx{^*}\gamma^o$ and prototypes $P^c$ of concept $c$ (see Eq.~\ref{eq:concept_response}); as distance measure the \emph{Mahalanobis distance} $d_{\text{mah}}(\cdot)$ is applied.
\begin{equation}
\small
\phi^c(\leftidx{^*}\gamma^o) = \frac{  \sum^{|P^c|}_{i=1} d_{\text{mah}}( \leftidx{^*}\gamma^o, p_i \in P^c )}{|P^c|}
\label{eq:concept_response}
\end{equation}

\section{Experimental Evaluation} \label{sec:experiment}
For evaluation purposes, we created a publicly available dataset, \emph{Object Shape Category Dataset}\footnote{\textbf{http://www.robotics.jacobs-university.de/datasets/2017-object-shape-category-dataset-v01/index.php}} (OSCD), that consists of about $66$ point cloud scans per category where each category contains multiple object instances. 
The 2.5D object point cloud scans of seven categories (see examples in Fig.~\ref{fig:ph_concept_extraction_samples}) are randomly split into a training/testing set with an average ratio of $75\%$/$25\%$ per category.

\subsection{Topological Filtration}
\label{sec:eval:filtration}

\begin{figure}
  \small
  \centering  
   \subfigure[Barcode(H$_0$)]{\label{fig:barcode}\includegraphics[height=0.41\linewidth]{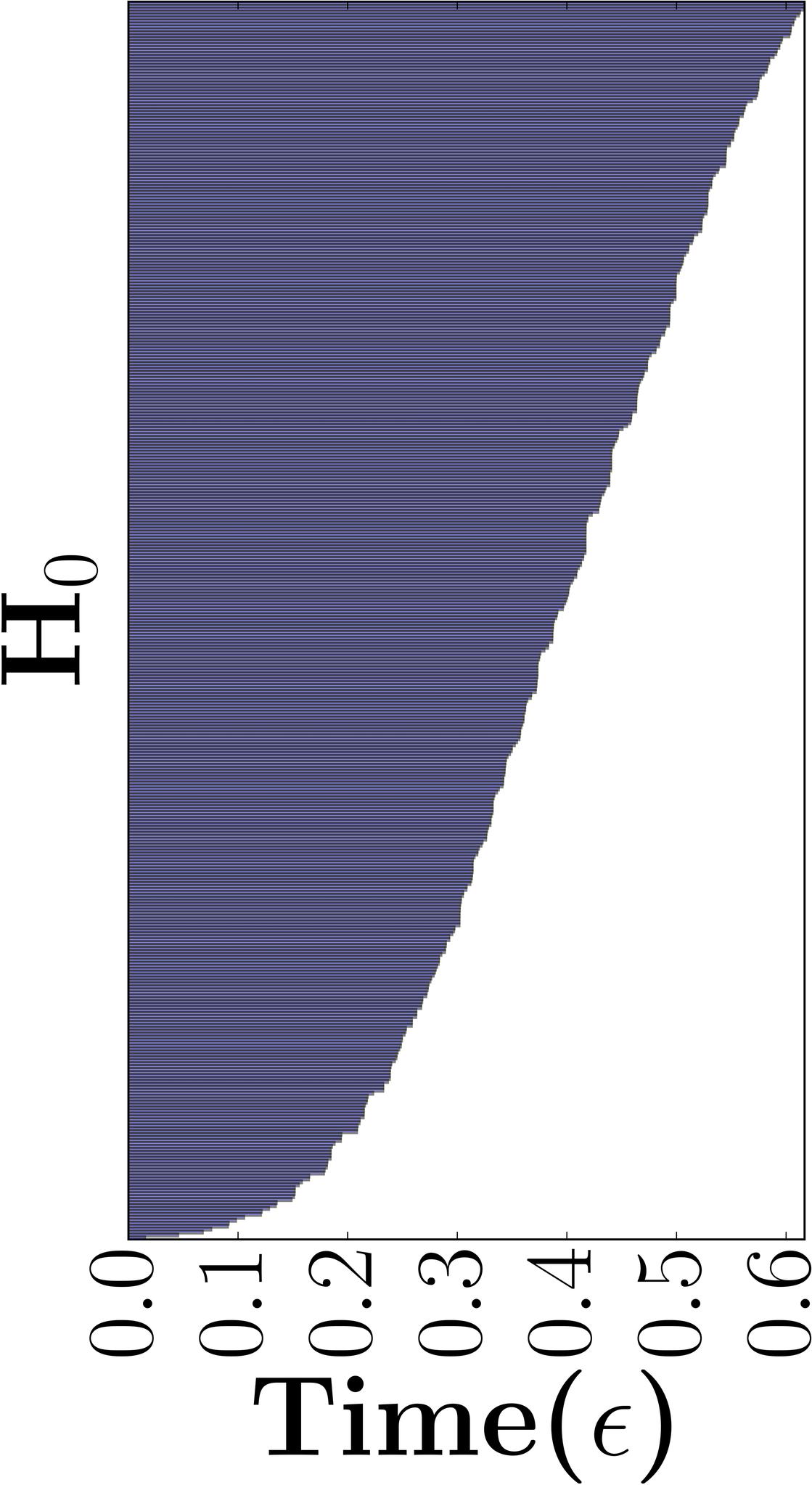}}
   \subfigure[Annexation]{\label{fig:annexation}\includegraphics[height=0.41\linewidth]{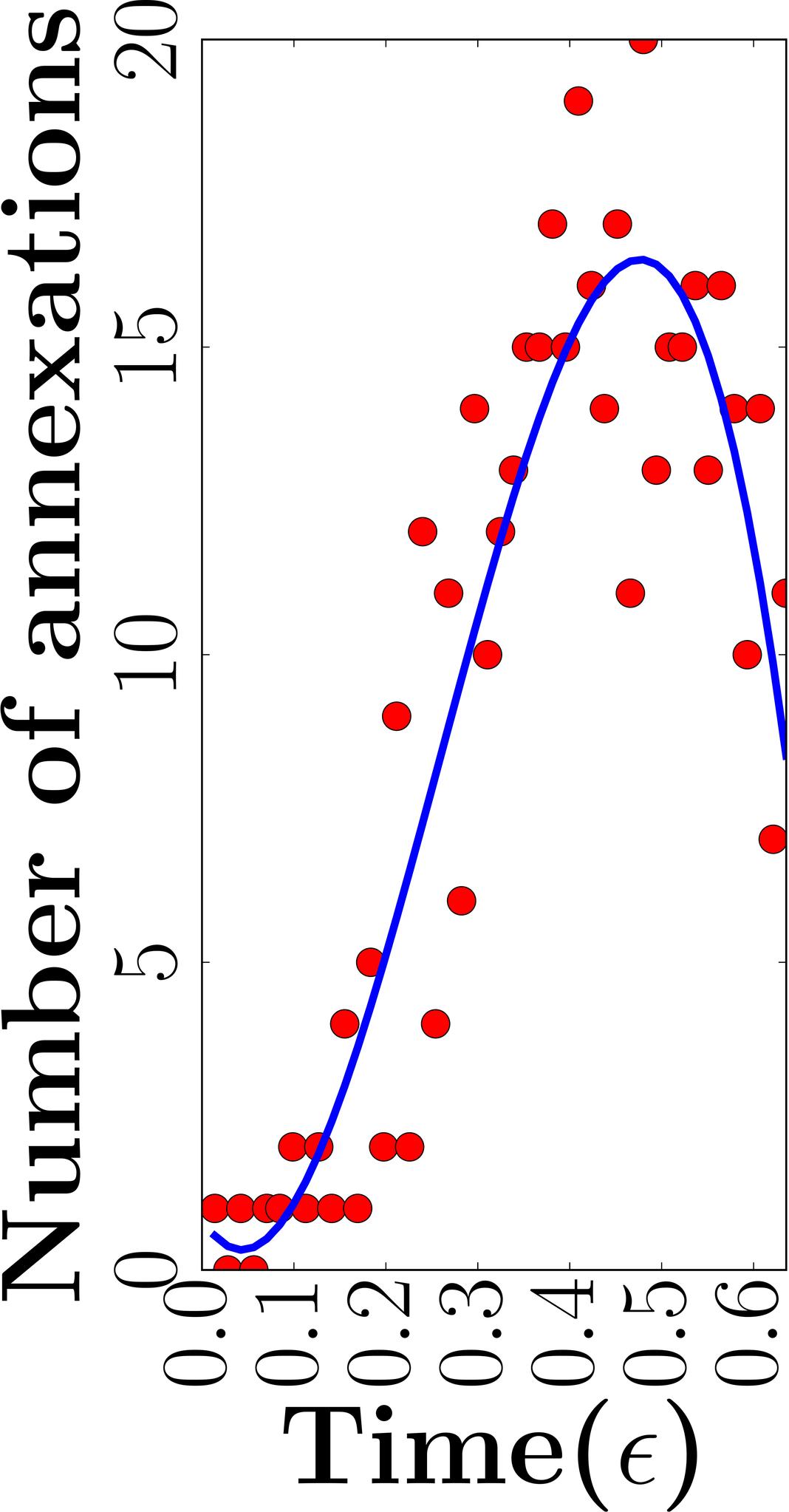}}
   \subfigure[Ranked concepts from Fig.~\ref{fig:ph_concept_extraction_samples}]{\label{fig:purity_proportion_result}\includegraphics[height=0.405\linewidth]{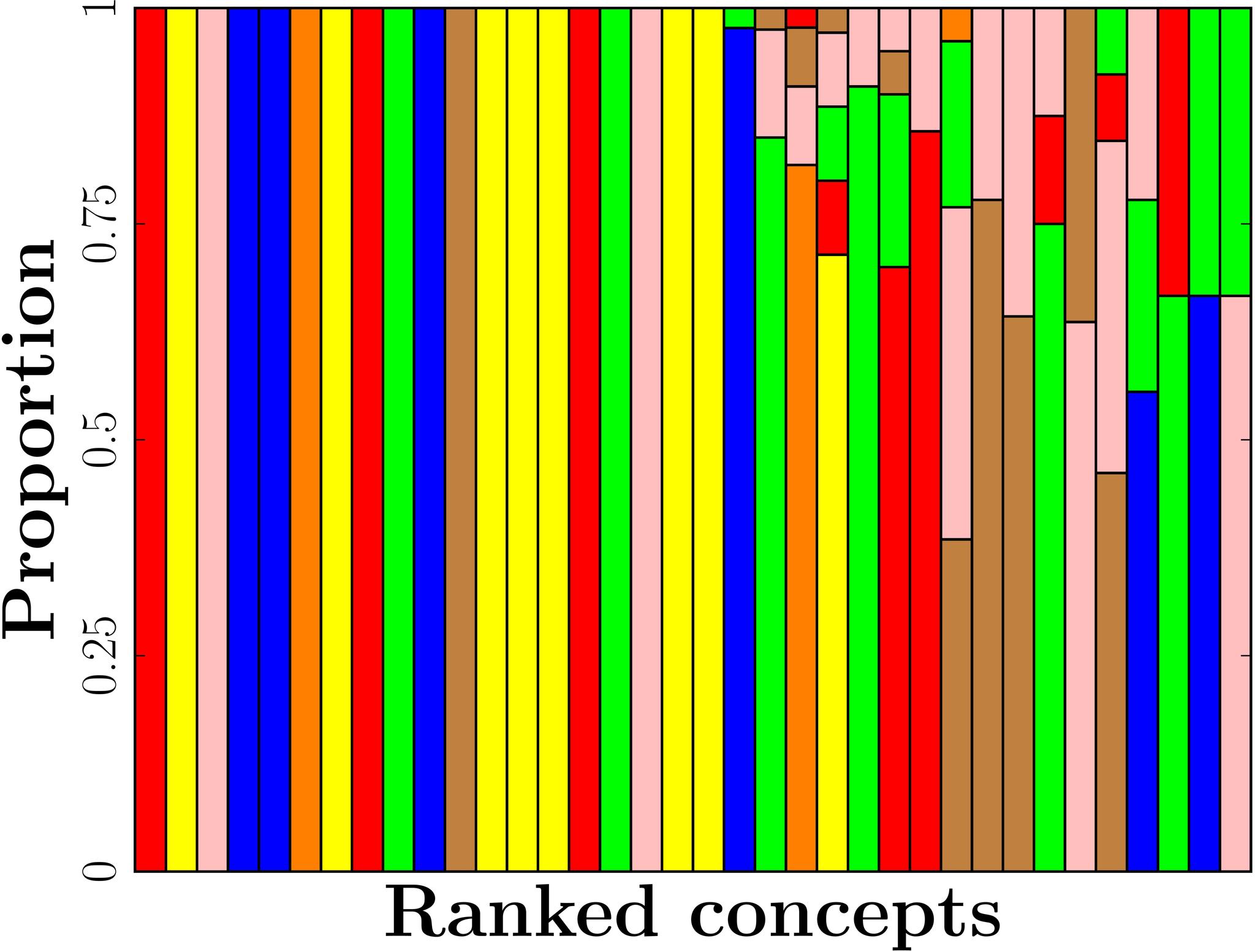}}
   \caption{Barcode of homology group 0 is shown in \subref{fig:barcode}. Number of annexations among Homology classes is shown \subref{fig:annexation}. In \subref{fig:purity_proportion_result} the proportional distribution of prototypes per concept. For visualization purpose, each proportion within a bar is colored with the corresponding label according to Fig.~\ref{fig:mst} and sorted in ascending order by $rs(\cdot)$, see Eq.~\ref{eq:rank_score}.}
   \label{fig:concept_generation_result}
\end{figure}

In the training phase, each training sample scan is propagated through  $\mathcal{HE}\mathrm{=}\{\mathcal{H}_1, \mathcal{H}_2,$ $...,$ $\mathcal{H}_n\}$, omitting any label-related information, i.e. each scan is unsupervisedly applied to the $\mathcal{HE}$; in our evaluation $n\mathrm{=}4$ has been heuristically determined -- a smaller $n$ may not allow $\mathcal{HE}$ to sufficiently discriminate the observed range of object shape variety.  
Afterwards, extracted stimuli vectors are fed to the filtration process (see Sec.~\ref{sec:desc_topo_analysis}).
Fig.~\ref{fig:association_graph} illustrates the filtration result of the stimuli vectors; the visualization does not reflect metric differences, it visualizes topological similarities among samples.
Already at this stage topological similarity can be observed w.r.t. the given category labels. 
Note that category labels are only associated to the prototypes for visualization purposes.
In Fig.~\ref{fig:barcode} the barcode is shown of the homology group 0.
At time $\epsilon_0$ all \emph{concept prototypes} -- depicted as bars -- are born. 
While the filtration progresses, more and more prototypes form larger homology classes that lead to the death (end of a bar) of prototypes which have been annexed.
As a result, only a single simplex at time $\epsilon_j$ survived the filtration (see Sec.~\ref{sec:topo_filration}). 
Moreover, Fig.~\ref{fig:annexation} shows only the number of annexation of homology classes over time.
It can be observed that the filtration reaches a global maximum of annexations at $\epsilon_{max}\mathrm{=}0.48$, i.e. the annexation of classes decreases even though $\epsilon$ reaches its maximum value.
It can be interpreted that the extracted homology classes after $\epsilon_{max}\mathrm{=}0.48$ are already discriminative by their persistence.

\subsection{Concept Learning}
The gradual filtration process as described in Sec.~\ref{sec:filt_PH}, allows to analyze the topological space at any filtration step. 
Each filtration step offers insights about the topology and relation among concept prototypes. 
Note that, the choice of a specific number of concepts and concept size depends on the objective of the application scenario. 

\subsubsection{Unsupervised Concept Selection}
\label{sec:unsup_concept_learning}
Using $\epsilon_{max}$ as indicator to stop the filtration process and subsequently selecting the existing homology classes at time $\epsilon_{max}$ as concepts, we receive in total $36$ concepts $\mathcal{C}$ (see Fig.~\ref{fig:ph_concept_extraction}) with a minimum concept size of $2$.
As discussed in Sec.~\ref{sec:shape_concept_extraction}, trivial extreme cases have to be considered: concept size of one, i.e. the number of concepts equals to the number of samples, and all samples belong to a single concept; concepts representing these cases are not considered.
To assess the quality of the extracted concepts we can make use of the human-annotated category labels which are associated to the prototypes (see Fig.~\ref{fig:ph_concept_extraction}).
Therein the purity of each concept can be interpreted as an external concept quality measure.
\emph{Purity} $pu(\cdot)$ is defined as the largest proportion in the distribution of prototypes of a category label, see Eq.~\ref{eq:purity}, where concept $c\in\mathcal{C}$ consists of a set of concept prototypes $P^c\mathrm{=}\{p_1, p_2, ...\}$ which are accordingly attributed with labels $Y^c\mathrm{=}\{y_1, y_2, ...\}$, i.e. $y_i\mathrm{=} \mathrm{retrieve\_label}(p_i)$, given the set of category labels $\mathcal{Y}$ of the dataset where $y_i\in\mathcal{Y}$.
\begin{equation}
\small
\small
pu(c) = \argmax_{y\in\mathcal{Y}} \frac{\sum^{|P^c|}_{i=1}\mathds{1}_y(y_i\in Y^c)}{\big|P^c\big|}
\label{eq:purity}
\end{equation}

Given the concepts inferred by $\epsilon_{max}$ as described in Sec.~\ref{sec:shape_concept_extraction} and shown in Fig.~\ref{fig:ph_concept_extraction}, it can be observed that connected components of different sizes are extracted which is caused by the shape heterogeneity of prototypes in $\mathcal{X}$. 
A large portion of the concepts is pure (see Eq.~\ref{eq:purity}), i.e. only prototypes of a specific category $y\in\mathcal{Y}$ are assigned to a concept $c\in\mathcal{C}$.
In Fig.~\ref{fig:purity_proportion_result} the resulting distribution of prototypes within a concept is illustrated.
Concepts are sorted in ascending order by the \emph{rank score} $rs(c)$ which considers the concept purity $pu(c)$ w.r.t. concept size $\big|P^c\big|$, see Eq.~\ref{eq:rank_score}. 
\begin{equation}
\small
rs(c) = \frac{\big|P^c\big|}{1 - pu(c) + \varepsilon},\text{\footnotesize where } \varepsilon \mbox{\ \footnotesize is a small constant\ } (0\mathrm{<} \varepsilon \mathrm{\ll} 1)
\label{eq:rank_score}
\end{equation}
While $57\%$ of the concepts are pure, other concepts show a lower purity, i.e. samples of different categories are assigned to a particular concept, however these categories show shape similarities like \emph{sack} and \emph{can} or \emph{plate} and \emph{box}. Furthermore, mean concept purity of $86.2\%$ is achieved.

Given the $36$ concepts, responses are extracted for each sample of the dataset, i.e. each sample object $o$ is represented by $\rho^o\mathrm{=}\{\phi^{1}( \leftidx{^*}\gamma^o), \phi^{2}(\leftidx{^*}\gamma^o),...\}$ ($|\rho^o|\mathrm{=}|\mathcal{C}|\mathrm{=}36$) and labeled with the corresponding dataset label. 
Accordingly a Support Vector Machine (SVM) is trained and evaluated, see Table~\ref{tab:unsupervised_comparison}.
\begin{table}%
 \small
\caption{Unsupervised concept selection: testing set (5 repetitions)}
\label{tab:unsupervised_comparison}
\centering
\footnotesize
\begin{tabular}{p{2.5cm}|p{0.25cm}p{0.25cm}p{0.20cm}p{0.35cm}p{0.15cm}p{0.75cm}p{0.5cm}}
Label: & sack & can & box & teddy & ball & amphora & plate\\ 
\hline
Mean error~(\%): &4.2& 6.5& 2.5& 8.8& 0& 10.4& 0\\
\end{tabular}
\begin{flushleft}
\end{flushleft}
\end{table}
Discriminative results have been obtained which allow to conclude the reasonability of the extracted concepts, e.g. shapes as \emph{ball}, \emph{plate} or \emph{box} show low cross-validation error, whereas appearance variety of categories caused by shape deformability and viewpoint changes, \emph{teddy} and \emph{amphora} may appear more ambiguous.

\subsubsection{Supervised Concept Selection}
\begin{figure}
  \small
  \centering  
   \subfigure[]{\label{fig:ph_concept_purity}\includegraphics[width=0.48\linewidth]{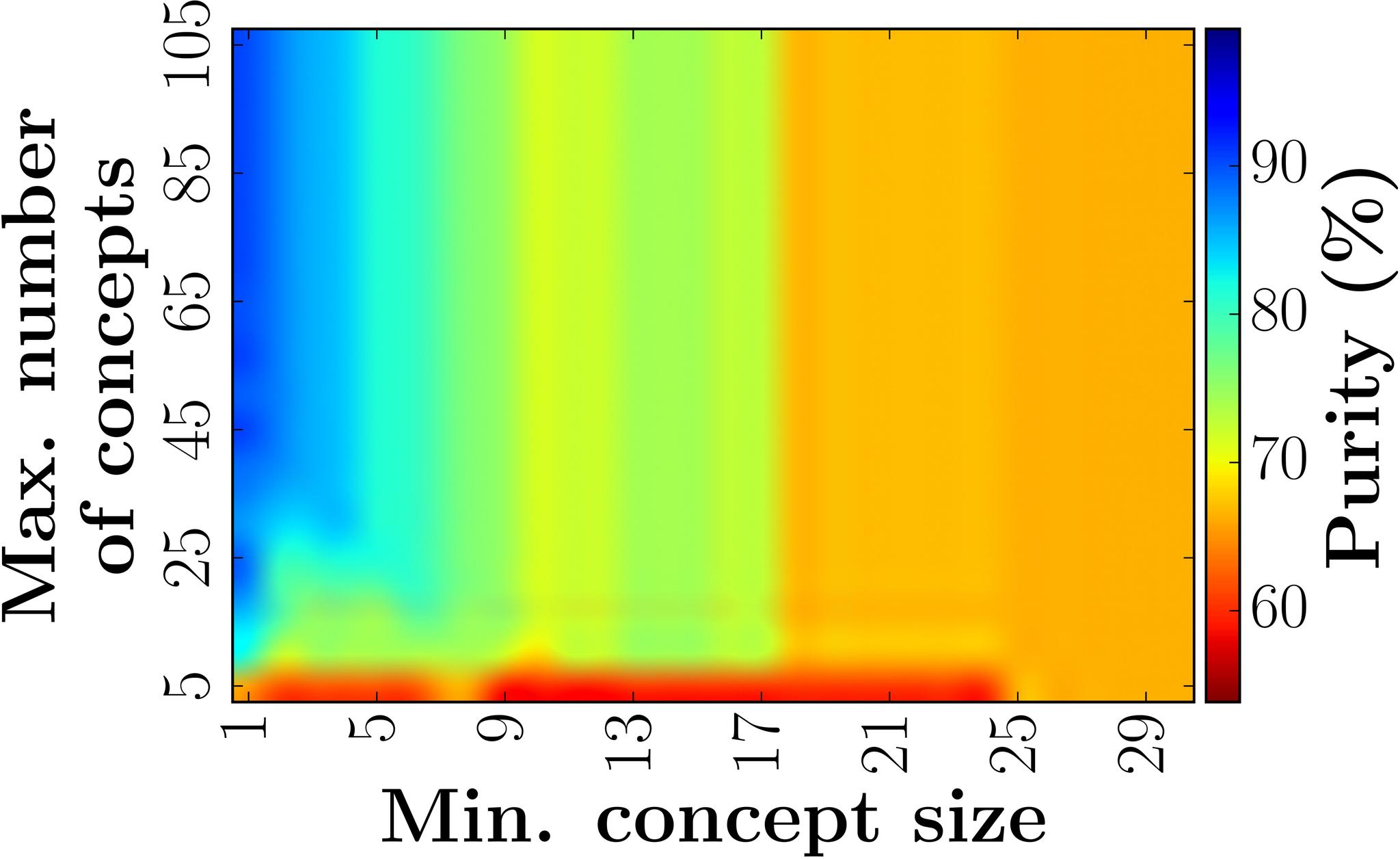}}
   \subfigure[]{\label{fig:sup_concept_error}\includegraphics[width=0.4775\linewidth]{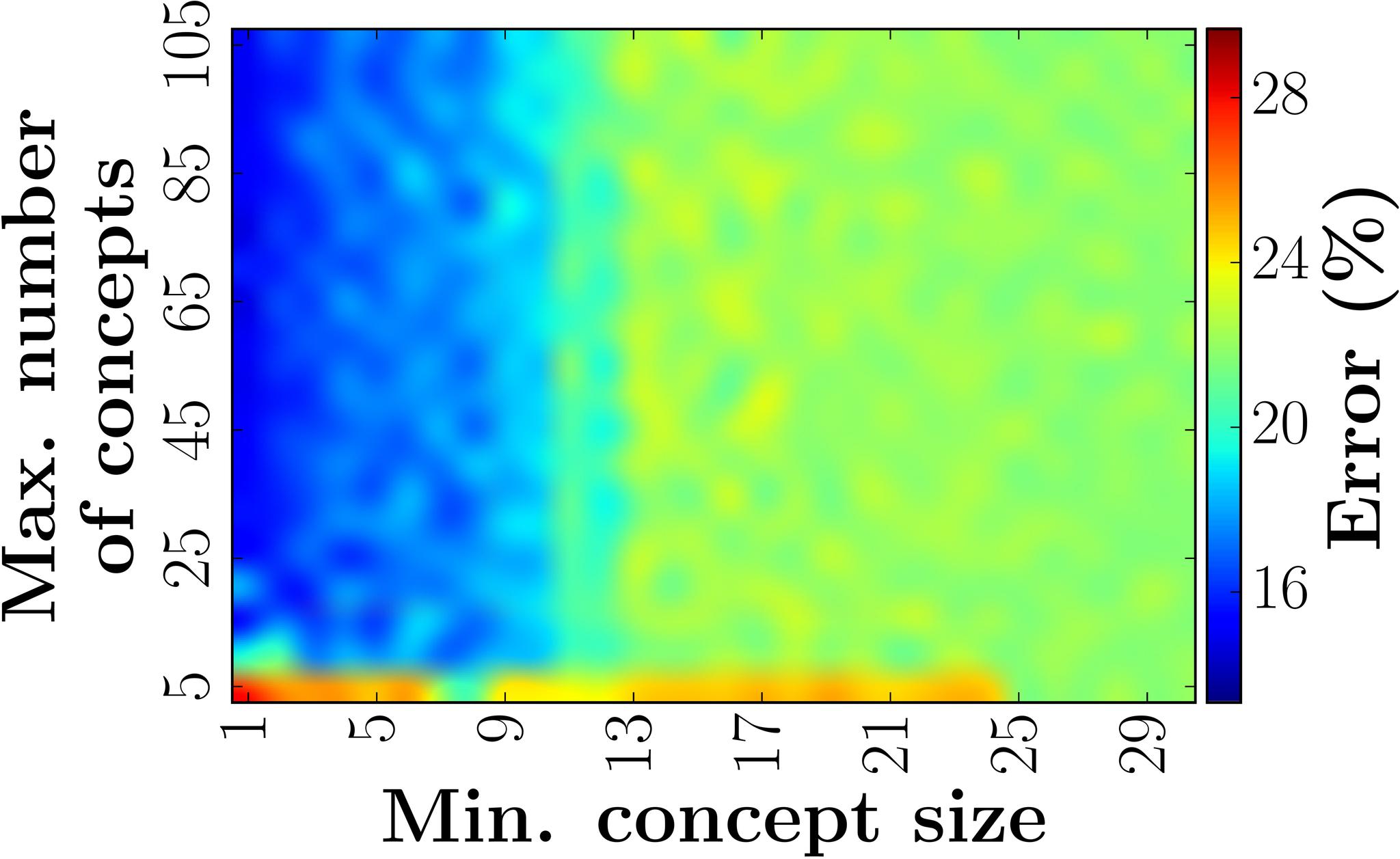}}
   \caption{Cross-validation result regarding concept purity \subref{fig:ph_concept_purity} and error \subref{fig:sup_concept_error} considering number and size of concepts.}
   \label{fig:concept_result}
\end{figure}

In contrast to Sec.~\ref{sec:unsup_concept_learning} where the filtration process is stopped in an \emph{unsupervised} manner by introducing $\epsilon_{max}$ and neglecting labels in order to identify reasonable concepts, in this section the decision is made in a \emph{supervised} manner.
Given the respective dataset labels for the concept responses, a cross-validation is performed over the parameter space (concept number and concept size) regarding the mean \emph{concept purity} and the \emph{classification error}, see Fig.~\ref{fig:concept_result}.
Note that, concepts are generated according to the parameters of the respective cross-validation step.
As in the filtration case, excluding the trivial cases as the minimum concept size of one and the number of concepts equals to the number of samples, only with $5$ concepts already a purity of $50\%$ per concept has been achieved.
The highest purity with $90.1\%$ is achieved with $68$ concepts and a minimum concept size of $2$, see Fig.~\ref{fig:ph_concept_purity}.
Regarding the mean classification error (see Fig.~\ref{fig:sup_concept_error}), concept responses are retrieved to train an SVM as performed in Sec.~\ref{sec:unsup_concept_learning}.
It is observable in Fig.~\ref{fig:concept_result} that higher concept purity can generally lead to a lower classification error which can be interpreted as, the feature quality (concept response) has increased by increasing concept purity.
The configuration with $68$ concepts and a minimum concept size of $2$ achieved a minimum cross-validation error of $14.79\%$ while reaching the highest purity with $90.1\%$.
Subsequently, as in Table~\ref{tab:supervised_comparison} shown %
\begin{table}%
\small
\caption{Supervised concept selection: testing set (5 repetitions)}
\label{tab:supervised_comparison}
\centering
\footnotesize
\begin{tabular}{p{2.5cm}|p{0.25cm}p{0.25cm}p{0.20cm}p{0.35cm}p{0.15cm}p{0.75cm}p{0.5cm}}
\footnotesize
Label: & sack & can & box & teddy & ball & amphora & plate\\ 
\hline
Mean error~(\%): &3.3& 6.8& 2.4& 8.9& 0& 10.7& 0\\
\end{tabular}
\begin{flushleft}
\end{flushleft}
\end{table}
similar classification results are achieved compared to the unsupervised concept selection shown in Table~\ref{tab:unsupervised_comparison} which is based on extracted concepts using $\epsilon_{max}$.
This outcome suggests that $\epsilon_{max}$ represents an indicator for stopping the filtration process and to subsequently extract concepts.

\subsection{Concept Discriminability with Alternative Datasets}
\label{sec:alter_db}
This experiment focuses on the generalization ability of the proposed approach.
Initially $\mathcal{HE}$ is trained \textbf{once} with the training set of our OSCD dataset.
Note that, this training process is unsupervised, i.e. $\mathcal{HE}$ is solely trained with instances in a label-agnostic manner. 
Subsequently, instances from the OSCD dataset are propagated through the $\mathcal{HE}$ (see Sec.~\ref{sec:spatial_topo_analysis}).
Based on the resulting stimuli vector of the propagation, concepts $\mathcal{C}$
are \textbf{once} generated (see Sec.~\ref{sec:desc_topo_analysis}).

\begin{figure}
	\scriptsize
	\centering
	\begin{minipage}[b]{.12\linewidth}	
    \begin{flushright} 
			\def \scaleImg{4.4} 
			\subfigure[]{\label{fig:eval:instance_variety0}\scalebox{\scaleImg}{\includegraphics[height=0.2\linewidth]{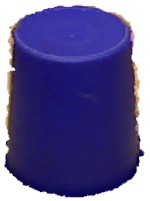}}}
			\subfigure[]{\label{fig:eval:instance_variety1}\scalebox{\scaleImg}{\includegraphics[height=0.17\linewidth]{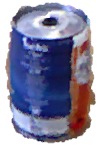}}}
			\subfigure[]{\label{fig:eval:instance_variety2}\scalebox{\scaleImg}{\includegraphics[height=0.06\linewidth]{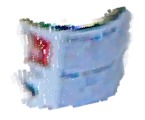}}}
			\subfigure[]{\label{fig:eval:instance_variety3}\scalebox{\scaleImg}{\includegraphics[height=0.09\linewidth]{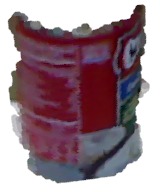}}}
			\subfigure[]{\label{fig:eval:instance_variety4}\scalebox{\scaleImg}{\includegraphics[height=0.11\linewidth]{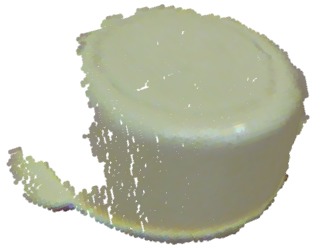}}}
			\subfigure[]{\label{fig:eval:instance_variety5}\scalebox{\scaleImg}{\includegraphics[height=0.12\linewidth]{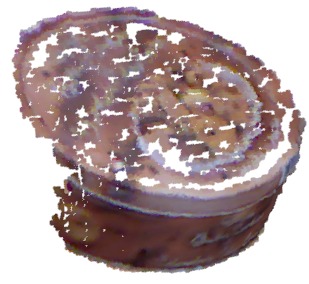}}}
		\end{flushright}
	\end{minipage} \hspace{0.5cm}
	\begin{minipage}[b]{.6\linewidth}
		\centering
		\subfigure[$\mathcal{CR}^2$ space (instance annotations scaled for visibility)]{\label{fig:unsup_tsne_sr}\includegraphics[width=0.99\linewidth]{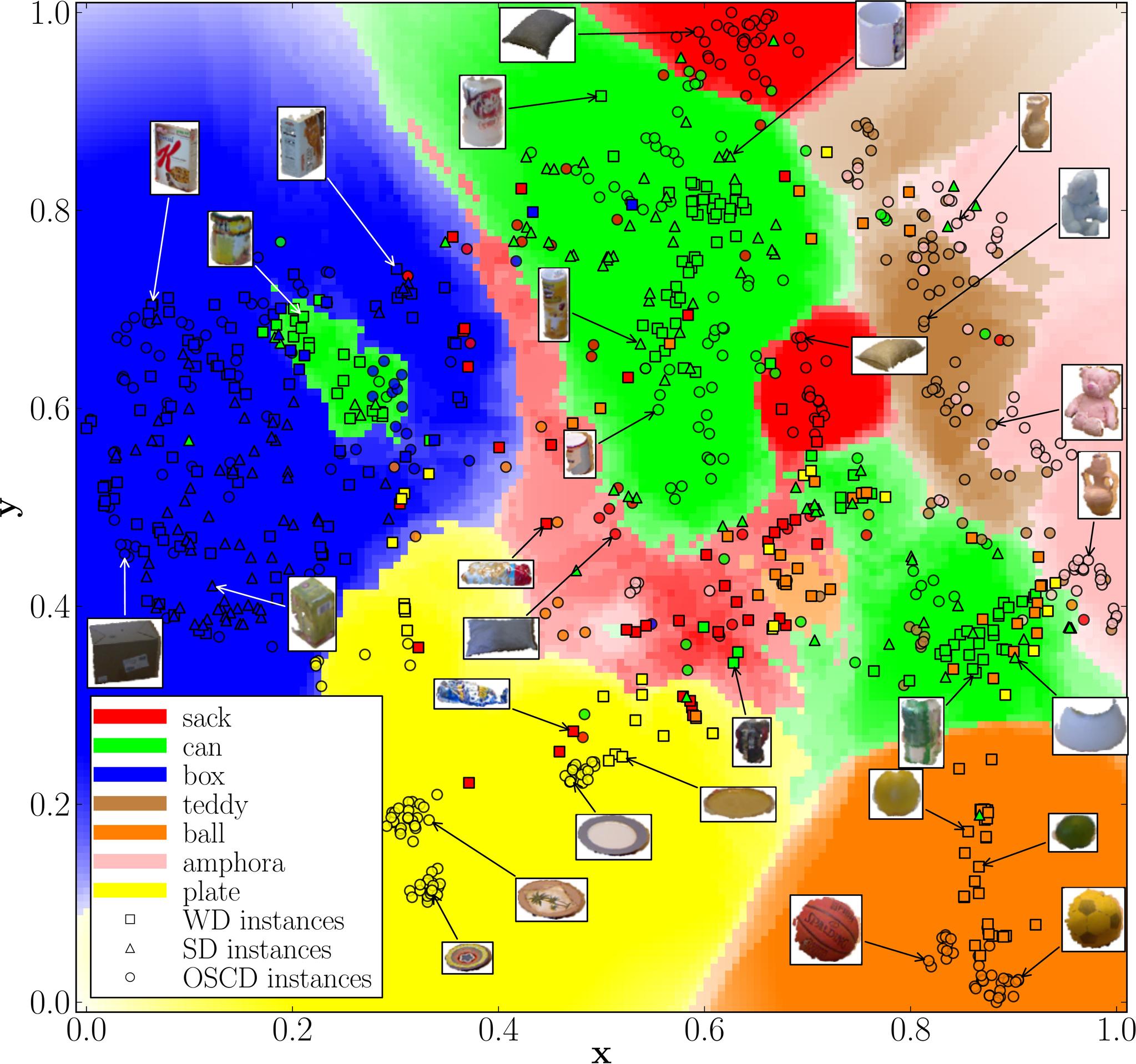}}
	\end{minipage}	
	\subfigure[Sample distribution of $\mathcal{CR}^2$ space. Note, for each instance of WD the $1^{st}$ to $5^{th}$ point cloud scans are selected of the first video sequence.]{\label{fig:tsne_instances_dist}
	\tiny
	\setlength{\tabcolsep}{0.12em}
	\begin{tabular}[b]{ l|| l |r|l| r |l |r ||r }
		Label& \multicolumn{2}{c|}{WD~\cite{5980382} scans\ \ \#} &  \multicolumn{2}{c|}{SD~\cite{6385661} scans \#}&  \multicolumn{2}{c||}{OSCD scans \ \ \ \ \ \ \ \ \ \ \ \ \ \ \ \ \ \ \ \ \ \#}&$\Sigma$\\ \hline
		\hline		
		\multirow{2}{*}{\emph{sack}} &  \emph{food bag} 1-8 & 40& & & \emph{sack} 0-56 (training set)&57&\multirow{2}{*}{115}\\
	 &  & & & &\emph{sack} 0-17 (testing set)&18&\\ \hline
		\multirow{2}{*}{\emph{can}}  &  \emph{food can} 1-14 &70&  \emph{learn} 33-44 &38& \emph{can} 0-59 (training set)&60&\multirow{2}{*}{259}\\
		&  \emph{soda can} 1-6 & 30&\emph{test} 31-42&42      & \emph{can} 0-18 (testing set) & 19&\\ \hline
				\multirow{2}{*}{\emph{box}} & \emph{cereal box} 1-5 & 25& \emph{learn} 0-16&38& \emph{box} 0-53  (training set)&54&\multirow{2}{*}{232}\\
		& \emph{food box} 1-12 & 60&            \emph{test} 0-15 &36& \emph{box} 0-18 (testing set) &19&\\ \hline
				\multirow{2}{*}{\emph{teddy}} &   &  & & &teddy 0-44 (training set) &45 &\multirow{2}{*}{59}\\
		 &   &  & & &teddy 0-13 (testing set) &14 &\\ \hline
			\multirow{3}{*}{\emph{ball}}&  \emph{ball} 1-7 & 35&     &&\emph{ball} 0-39  (training set)&40&\multirow{3}{*}{125}\\
		&  \emph{lime} 1-4 & 20&                 && \emph{ball} 0-9 (testing set)&10&\\
		&  \emph{orange} 1-4 & 20&                &&  &&\\ \hline
			\multirow{2}{*}{\emph{amphora}}&   &  & & &\emph{amphora} 0-47 (training set)& 48&\multirow{2}{*}{62}\\
		&   &  & & &\emph{amphora} 0-13 (testing set)& 14\\ \hline
	\multirow{2}{*}{\emph{plate}}& \emph{plate} 1-7 &35&      && \emph{plate} 0-49 (training set)&50&\multirow{2}{*}{105}\\ 
		& &&      && \emph{plate} 0-19 (testing set)&20 \\ \hline \hline
		$\Sigma$&  \multicolumn{1}{c|}{-} & 335 &     \multicolumn{1}{c|}{-}      &154&       \multicolumn{1}{c|}{-}  & 468 & 957
	\end{tabular} 
} 
	\subfigure[]{\label{fig:eval:oscd_cfmat:dist}\includegraphics[width=0.4\linewidth]{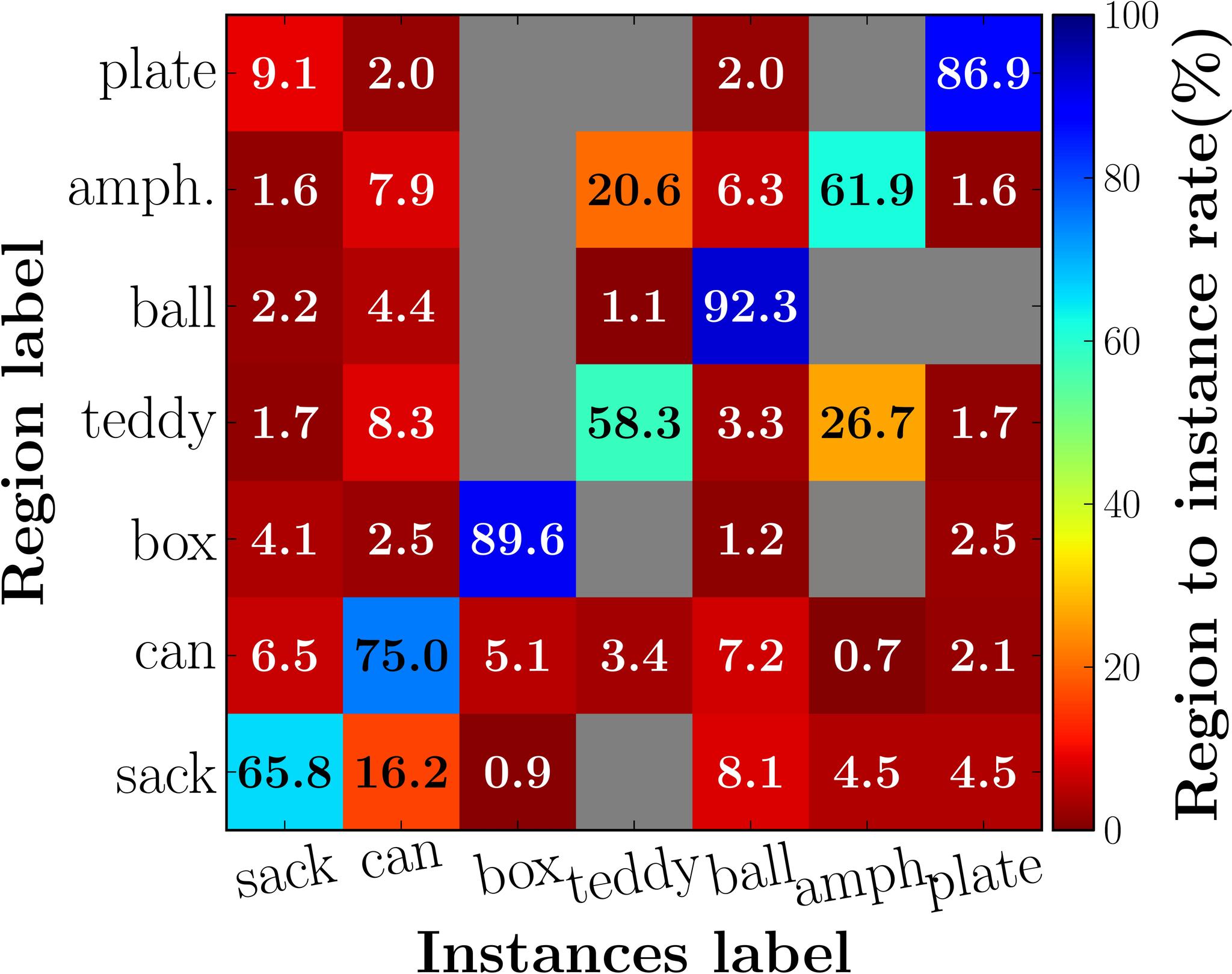}}
	\subfigure[]{\label{fig:eval:oscd_cfmat:assigm}\includegraphics[width=0.4\linewidth]{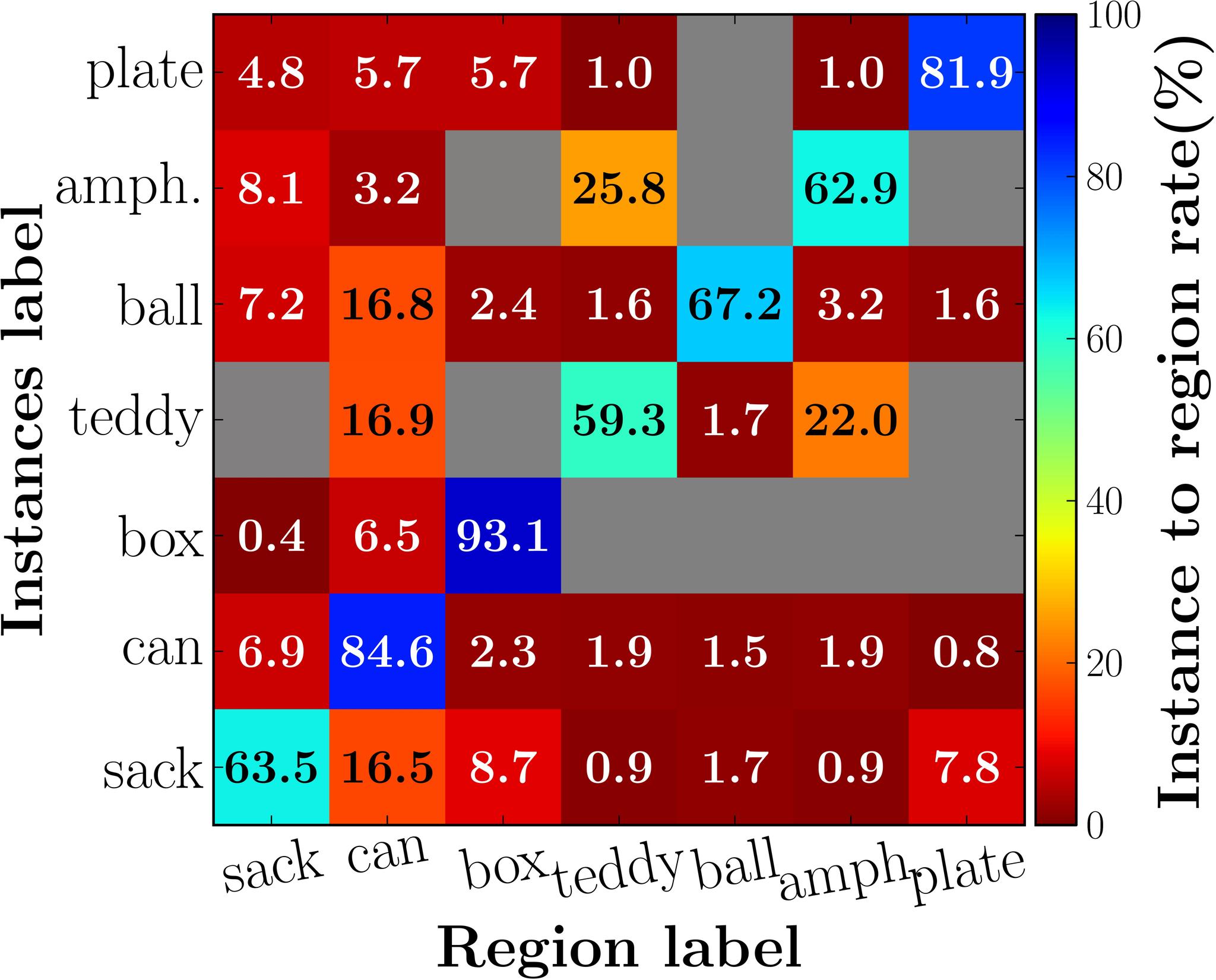}}
	\caption{In \subref{fig:eval:instance_variety0}-\subref{fig:eval:instance_variety5} appearance variations of cylindric samples from different datasets: \subref{fig:eval:instance_variety0}, \subref{fig:eval:instance_variety1} show \emph{can}~\emph{0} and \emph{56} of OSCD-training set; \subref{fig:eval:instance_variety2} and \subref{fig:eval:instance_variety3} show \emph{food\_can\_1\_1\_1} and \emph{food\_can\_14\_1\_1} of WD dataset~\cite{5980382};  \subref{fig:eval:instance_variety4}, \subref{fig:eval:instance_variety5} show cylindrical instances from scenes \emph{learn 34} and \emph{test 42} of the SD dataset~\cite{6385661}.
	In \subref{fig:unsup_tsne_sr}, $\mathcal{CR}^2$ is shown with instances \subref{fig:tsne_instances_dist} from WD, SD and OSCD dataset.
	Accordingly, the distribution is shown of instances within \emph{a region} \subref{fig:eval:oscd_cfmat:dist} and the assignment of instances to \emph{particular regions} \subref{fig:eval:oscd_cfmat:assigm}.}
	\label{fig:unsup_tsne}
\end{figure}

Given the previously trained $\mathcal{HE}$ model and the generated concepts $\mathcal{C}$, in the following we evaluate the discriminability of the generated concepts with instances of the OSCD, \emph{Washington RGB-D Object Dataset}~\cite{5980382} (WD) and \emph{Object Segmentation Database}~\cite{6385661} (SD) (Fig.~\ref{fig:tsne_instances_dist}); note that all three datasets are sampled from \emph{different distributions} as illustrated in Fig.~\ref{fig:eval:instance_variety0}-\subref{fig:eval:instance_variety5}. 
In order to analyze the spectrum of responses for these dataset objects, each object $o$ is initially represented with as graph of segments $g^o$ (see Sec.\ref{sec:obj_seg_dict}) and applied to the two-step procedure: \textbf{1)} propagate $g^o$ through $\mathcal{HE}$ to generate a stimuli vector  $\leftidx{^*}\gamma^o$ (see Sec.~\ref{sec:stimuli_generation}); \textbf{2)} compute for each concept $c\in \mathcal{C}$ the response with $\phi^c(\leftidx{^*}\gamma^o)$ (see Eq.~\ref{eq:concept_response} in Sec.~\ref{sec:shape_concept_inference}).
As a result an object $o$ is represented by the set of concept responses $\rho^o\mathrm{=}\{ \phi^1(\leftidx{^*}\gamma^o),\phi^2(\leftidx{^*}\gamma^o),...\}$ ($|\rho^o|\mathrm{=}|\mathcal{C}|\mathrm{=}36$, see Fig.~\ref{fig:ph_concept_extraction}).

Consequently, a $|\mathcal{C}|$-dimensional space of concept responses $\mathcal{CR}^{|\mathcal{C}|}$ is created.
The generalization capability can be investigated by $\mathcal{CR}^{|\mathcal{C}|}$ which allows to observe relations and similarities among sample objects.
To visualize and reason about the $|\mathcal{C}|$-dimensional $\mathcal{CR}^{|\mathcal{C}|}$ space, the \emph{t-SNE}~\cite{ictdbid:2777} embedding technique is applied to reduce the dimensionality to two; this 2D space we denote as $\mathcal{CR}^2$.
The embedding is performed in an unsupervised manner, i.e. label-agnostic.
As a result instances are projected to a two-dimensional $\mathcal{CR}^2$ space, from WD, SD and OSCD dataset, see Fig.~\ref{fig:unsup_tsne}.
For illustration, regions in $\mathcal{CR}^2$ are colored according to their dedication to a certain label (see Fig.~\ref{fig:unsup_tsne_sr}) by exploiting the projected instances as anchor points in space. 
Therefore, a uniform grid is created within the 2D $\mathcal{CR}^2$ space; for each cell in the grid the \emph{k}-nearest instances are determined (e.g. \emph{k}$\mathrm{=}$5\% of total number of instances); then the majority label of the \emph{k} instances is determined and the cell is colored according to the majority label; each
cell is weighted and visually depicted in form of
cell opacity. 
The weight represents the observed proportion of the \emph{k} instances associated to majority label which is depicted in an interval $[0, 1]$ from low to high proportion [low: transparent (white)$\mathrm{=}$0, high: opaque (solid majority label color)$\mathrm{=}$1].

The continuous space $\mathcal{CR}^2$ shown in Fig.~\ref{fig:unsup_tsne_sr} allows to learn regional characteristics and relations among locations in $\mathcal{CR}^2$ and instances of the three datasets.
A main observation is that instances from different datasets are propagated through the $\mathcal{HE}$ and the resulting concept responses show \emph{coherency} regarding shape appearance: instances of all evaluated datasets together can form interrelated and coherent groups, see uniformly colored regions in Fig.~\ref{fig:unsup_tsne_sr}.
This is also reflected in Fig.~\ref{fig:eval:oscd_cfmat:dist} and \subref{fig:eval:oscd_cfmat:assigm} that illustrate the distribution of instances in $\mathcal{CR}^2$ space.
Instances labeled as \emph{can}, \emph{box}, \emph{ball}, \emph{amphora}, \emph{plate} form distinct regions whereas deformable instances like \emph{sack} and \emph{teddy} lead to more scatter. 
However \emph{teddies} are still represented as a connected region and regions dedicated to \emph{sack} are located at transitions to other labeled regions, e.g. \emph{can} to \emph{plate}, \emph{can} to \emph{box} or \emph{can} to \emph{teddy}.
This observation can be explained that \emph{sacks} can be interpreted as an intermediate shape, e.g. between a \emph{box} and a \emph{can} in $\mathcal{CR}^2$ space due to their roundish, bulgy or cylindric appearance depending on viewpoint and deformation.
  
Note that, in the context of \emph{Cognitive Science}, specifically in the field of representation architectures, $\mathcal{CR}$ can also be interpreted as a \emph{Conceptual Space}~\cite{zenker2015} where points (prototypes) in space represent multidimensional vectors of \emph{stimuli}, and regions in space \emph{concepts}.

\section{Conclusion}
\label{sec:conclusion}

We proposed an unsupervised abstraction process, from 3D point clouds to semantically meaningful concepts of shape commonalities, that is applicable in various robotic areas (see Sec.~\ref{sec:intro}).
The proposed Shape Motif Hierarchy Ensemble encodes object segment compositions in a hierarchical symbolic manner.
Inspired by the concept of Persistent Homology, stimuli generated by the ensemble are filtered in a gradual manner to reveal topological structures.
The filtration leads to stimuli groups which can be interpreted as shape concepts that reflect \emph{commonalities} of shape appearances.
In experiments, commonalities are revealed and the generalization capability is shown by introducing unknown samples of external datasets. %
These concepts which are learned in an unsupervised (label-agnostic) fashion, have shown associations to human-annotated shape categories and that they can be used as features to train a supervised classifier for shape reasoning purposes such as shape category recognition.

\bibliographystyle{IEEEtran}

\end{document}